\documentclass[10pt,journal,compsoc]{IEEEtran}
\usepackage{amsmath,amsfonts}
\usepackage{algorithmic}
\usepackage{algorithm}
\usepackage{array}
\usepackage{subfig}
\usepackage{textcomp}
\usepackage{stfloats}
\usepackage{url}
\usepackage{verbatim}
\usepackage{graphicx}
\usepackage{cite}
\usepackage[pagebackref=true,breaklinks=true,letterpaper=true,colorlinks,bookmarks=false]{hyperref}

\usepackage{makecell}
\usepackage{boldline}
\usepackage{algorithm, algorithmic}
\usepackage{multirow}  
\usepackage{enumitem}
\usepackage{breqn}
\usepackage{cleveref} % for cref
\usepackage{wrapfig}
\usepackage{caption} % for subfigure
\usepackage{booktabs}
\usepackage{amsthm} % required for normal theorem definition.
\newtheorem{thm}{Theorem} 

\usepackage[table, x11names]{xcolor}
\definecolor{Gray}{gray}{0.93}

\crefname{section}{Sec.}{Secs.}
\Crefname{section}{Section}{Sections}
\Crefname{table}{Table}{Tables}
\crefname{table}{Tab.}{Tabs.}
\crefname{figure}{Fig.}{Figs.} 
\crefname{equation}{Eq.}{Eqs.}
\crefname{thm}{Thm.}{Thm.}

\newcommand{\etal}{\textit{et al.\ }}
\newcommand{\ie}{\textit{i.e.}}
\newcommand{\eg}{\textit{e.g.}}
\usepackage{ragged2e}

\begin{document}

\title{Exploring Flat Minima for Domain Generalization with Large Learning Rates}

% \author{IEEE Publication Technology,~\IEEEmembership{Staff,~IEEE,}
% <-this % stops a space
% \thanks{}% <-this % stops a space
% \thanks{Manuscript received  .}}

\author{
  Jian Zhang,
  Lei Qi$^*$,
  Yinghuan Shi$^*$,
  Yang Gao
  \thanks{}% <-this % stops a space
  \thanks{Jian Zhang, Yinghuan Shi, and Yang Gao are with the State Key Laboratory for Novel Software Technology and National Institute of Healthcare Data Science, Nanjing University, Nanjing 210023, China (e-mail: zhangjian7369@smail.nju.edu.cn; syh@nju.edu.cn; gaoy@nju.edu.cn).}
  \thanks{Lei Qi is with the School of Computer Science and Engineering, Southeast University, Nanjing 211189, China (e-mail: qilei@seu.edu.cn).}
  \thanks{
    $^*$Corresponding authors: Yinghuan Shi and Lei Qi.
  }
}

% The paper headers
% \markboth{IEEE TRANSACTIONS ON KNOWLEDGE AND DATA ENGINEERING }{Zhang \MakeLowercase{\textit{et al.}}: Exploring Flat Minima for Domain Generalization with Large Learning Rates}

% \IEEEpubid{0000--0000/00\$00.00~\copyright~2021 IEEE}
% Remember, if you use this you must call \IEEEpubidadjcol in the second
% column for its text to clear the IEEEpubid mark.

% Current methods to identify flat minima mainly involve techniques such as weight averaging or weight perturbation. 
% However, the gradient estimation of weight perturbation becomes inaccurate when the perturbation strength is high while traditional stochastic weight averaging requires diversity among weights.
% we propose a novel method to identify flat minima from the perspective of the learning rate. W
%  with a greater range of weight variations compared to a smaller learning rate
\IEEEtitleabstractindextext{
  \begin{abstract}
    \justifying
    Domain Generalization (DG) aims to generalize to arbitrary unseen domains. A promising approach to improve model generalization in DG is the identification of flat minima.
    One typical method for this task is SWAD, which involves averaging weights along the training trajectory. However, the success of weight averaging depends on the diversity of weights, which is limited when training with a small learning rate.
    Instead, we observe that leveraging a large learning rate can simultaneously promote weight diversity and facilitate the identification of flat regions in the loss landscape.
    However, employing a large learning rate suffers from the convergence problem, which cannot be resolved by simply averaging the training weights.
    To address this issue, we introduce a training strategy called Lookahead which involves the weight interpolation, instead of average, between fast and slow weights. The fast weight explores the weight space with a large learning rate, which is not converged while the slow weight interpolates with it to ensure the convergence.
    Besides, weight interpolation also helps identify flat minima by implicitly optimizing the local entropy loss that measures flatness.
    To further prevent overfitting during training, we propose two variants to regularize the training weight with weighted averaged weight or with accumulated history weight.
    Taking advantage of this new perspective, our methods achieve state-of-the-art performance on both classification and semantic segmentation domain generalization benchmarks. The code is available at \url{https://github.com/koncle/DG-with-Large-LR}.
  \end{abstract}
  \begin{IEEEkeywords}
    Domain Shift, Domain Generalization, Flat Minima, Large Learning Rate
  \end{IEEEkeywords}
}

\maketitle

\section{Introduction}

\IEEEPARstart{D}{omain} shifts in the real world, \ie, the training data's distribution is different from that of the test data, pose a great challenge to the independent identical distribution assumption in traditional supervised learning.
Traditional domain adaptation (DA) methods~\cite{venkateswara2017deep,saito2019semi,li2021faster,wu2020iterative} are designed to deal with this kind of domain shift during the test, which collects labeled source and target domain data to train a model that performs well in the target domain.
  {To improve the label efficiency, unsupervised domain adaptation (UDA)~\cite{xiao2021dynamic,yue2021transporting,kouw2019review} is proposed to transfer the source knowledge to the target domain with only unlabeled target data.}
However, this approach faces a significant problem in that the target domain is known in advance, which is not practical in the real world where the environment may change according to space or time. Besides, the cost of continuously collecting target domain data and retraining the deployed model for the dynamic environment is also unbearable. Therefore, Domain Generalization (DG)~\cite{muandet2013domain,wang2022generalizing,yuan2023collaborative} is proposed as a more practical setting that addresses this issue by generalizing the source data trained model to any unseen domain without retraining with target domain data. Through the training of a highly generalizable model, the need for a large number of human resources can be significantly reduced.

\begin{figure}[t]
  \center
  \includegraphics[width=0.9\linewidth]{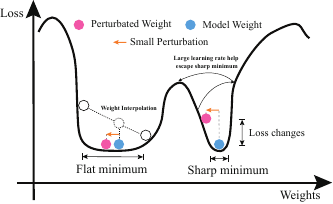}
  \caption{The comparison of the sharp and flat minimum and illustration of two key elements of our method (\ie, large learning rate and weight interpolation).}
  \label{fig_intro}
  % \vspace{-5pt}
\end{figure}

Due to its promising practical applications, numerous methods~\cite{li2018learning,dou2019domain,zhang2022mvdg,xu2021fourier} in DG have been proposed to improve generalization.
% There is a critical element in domain generalization that plays a critical role in improving generalization performance: preventing overfitting to the source domains.
Preventing overfitting to the source domains is the critical element in designing these methods.
To achieve this goal, methods that focus on different aspects (\eg, data augmentation at data or feature level~\cite{jin2020style,xu2021fourier}, domain-invariance learning~\cite{bousmalis2016domain, chen2021style,li2018domain,deng2020representation} at feature level) of the training procedure have been designed.
However, most of these methods, when evaluated using a fair DomainBed benchmark~\cite{gulrajani2020search}, even fail to outperform the performance of a simple baseline, Empirical Risk Minimization (ERM), which simply aggregates all training data to train the model without additional techniques, indicating that these methods only achieve partial generalizability that is specific to the training datasets. Consequently, when confronted with a more fair and diverse benchmark~\cite{gulrajani2020search}, they typically struggle to generalize effectively.
Despite being recognized as a strong baseline for Domain Generalization (DG), {just like the aforementioned methods}, ERM still inevitably suffers from performance degradation, which is caused by the sharp minimum in the loss landscape it converges to.
These sharp minima result in poorer generalization of the trained model compared to flat minima~\cite{cha2021swad}.
As illustrated in \cref{fig_intro}, models located in these sharp minima exhibit high sensitivity to small perturbations in the weight space. Additionally, since the changes in the weight space could potentially reflect the changes (\eg, the domain shift during the test) in the data space~\cite{zhang2022ga}, models converging to the sharp minimum lack robustness when confronted with domain shifts in unseen test data.
In contrast, adhering to the principles of the Minimum Description Length (MDL), flat minima possess the advantage of being representable with fewer bits, reducing the risk of overfitting to the source data that demands more bits.
Therefore, it is crucial to identify flat minima during training to enhance the model's generalization ability.

To better identify flat minima, researchers introduced the Sharpness-aware Minimization (SAM) \cite{foret2020sharpness} that searches for the flat area by pushing the training model away from the local sharp area, which can effectively escape from the sharp minima around the neighborhood. In contrast, prior techniques in domain generalization, such as SWAD \cite{cha2021swad} and SMA~\cite{arpit2021ensemble}, take a different route.
  {These methods leverage the observation that averaging the weights from the periphery of the loss landscape can yield weights approximating the flat regions near the center \cite{izmailov2018averaging}. They achieve this by averaging all training weights along the trajectory of the training process, which implicitly shifts the weights towards the central, flatter regions of the loss landscape. However, the effectiveness of weight averaging is closely related to the diversity of weights~\cite{rame2022diverse}, which is challenging when training with a small learning rate.
    Conversely, we have observed that utilizing a large learning rate can produce diverse weights at both feature and prediction levels compared to a small learning rate.
    Besides, a large learning rate also helps identify flat areas, allowing the training model to have a greater chance to jump out of the local sharp minimum and stay in the flatter regions, as depicted in \cref{fig_intro}. When a small learning rate is employed, the training model is more susceptible to getting trapped in sharp local minima, making it challenging to escape. In contrast, due to the narrow loss landscape typically associated with the sharp minimum, a large learning rate enables the model to easily move away from these minima and explore wider and flatter minima~\cite{seong2018towards}.
    Therefore, in this paper, we propose to utilize a large learning rate to both improve the weight diversity and identify flat minima.
  }

  {However, a mere increase in the learning rate in the standard training procedure can result in significant performance degradation because of the difficulty in convergence. Besides, weight averaging cannot address this issue since it does not guarantee convergence.
    To deal with the above problem, instead of averaging the weights, we propose to adopt the weight interpolation operation by introducing the training strategy with fast and slow weights, named Lookahead~\cite{zhang2019lookahead}. It first explores the weight space with fast weight in a large learning rate and then different from weight averaging, it interpolates the fast and slow weight to achieve a balance between optimization and generalization~\cite{zhou2021towards}.
    We prove that this interpolation operation can effectively ease the optimization problem of the large learning rate while preserving its benefits in finding flat minima and producing diverse weights.
    Besides, several studies~\cite{arpit2021ensemble,izmailov2018averaging,frankle2020linear} also demonstrate that a model located within the interpolated area of two weights that share the same training trajectory tends to stay in a flat region with lower loss, as shown in \cref{fig_intro}. Therefore, interpolation operation can also help identify flat minima and we prove that it also implicitly maximizes the local entropy loss~\cite{chaudhari2019entropy} that measures flatness.
    % of the current weight's neighborhood.
  }

Consequently, by incorporating the large learning rate and the interpolation operation during training, Lookahead can effectively identify the flat area and achieve good performance. However, it may still inevitably suffer from the overfitting issue during training where the training loss decreases and validation loss increases.
  {To alleviate this issue}, we propose two different strategies to regularize the training procedure. The first \textit{AvgLookahead} performs weight averaging along the training trajectory of fast weight to stabilize training and reduce the too-extreme exploration with the large learning rate, while the second \textit{RegLookahead} regularizes the training weight with the averaged history weights to regularize exploration.
We conduct the comparison on two widely employed classification and semantic segmentation benchmarks and our proposed methods achieve state-of-the-art performance.
% Furthermore, by comparing Lookahead to previous methods (\ie, SAM~\cite{foret2020sharpness} and SWAD~\cite{cha2021swad}), we discover several connections between these methods, which provide valuable insights for further research on the flat minimum searching algorithms.

Our contributions could be summarized as follows:
\begin{itemize} %[noitemsep,topsep=0pt]
  %connect meta-learning to flatness and provide two key elements of its success in finding flat minima.
  \item We provide a novel perspective to identify flat minima with a large learning rate in Domain Generalization.
  \item We theoretically and empirically validate the effectiveness of finding flat minima with a large learning rate and weight interpolation.
  \item To further alleviate overfitting, we propose two effective variants of Lookahead for regularization.
  \item Experiments on classification and segmentation benchmarks validate the effectiveness of our methods.
\end{itemize}

\section{Related Work}

\textbf{Domain generalization} (DG) has attracted great attention recently due to its ability to generalize to previously unseen domains while solely relying on knowledge from source domains~\cite{muandet2013domain}. Although the objective of domain generalization is to achieve robust generalization across novel domains, the training phase often involves only a limited number of available domains.  This inherent contradiction gives rise to a prominent challenge in DG: overfitting. Several strategies have been devised to combat overfitting, such as learning domain-invariant features~\cite{li2018deep,zhao2020domain,rahman2019correlation}, employing data augmentation techniques~\cite{yue2019domain,zhou2020deep,xu2020robust,zhou2021mixstyle,xu2021fourier}, and implementing various regularization methods~\cite{li2018learning,li2019episodic,li2019feature,cha2021swad}.

Since overfitting implies that a model learns domain-specific information, traditional methods~\cite{li2018deep,zhao2020domain,rahman2019correlation,guan2023rfdg} try to learn domain-invariant features that possess robust generalization capabilities across unfamiliar domains. Li \etal~\cite{li2018deep} learn the conditional domain-invariant features to ensure the invariant mapping across all domains, Rahman \etal~\cite{rahman2019correlation} learn the domain-invariant features by aligning the correlation matrix of learned features. Zhao \etal~\cite{zhao2020domain} introduces entropy minimization term to ensure the conditional invariance of data and label relationship across domains.

As commonly employed in traditional tasks, an intuitive way to overcome the overfitting problem is to augment the scarce source data with diverse styles to enhance the shape recognition of the trained model. In practice, augmentation has also been proven to be one of the most successful techniques for domain generalization. The augmentation in DG can take place either at the data level (\eg, generate new images)~\cite{chen2021style,guo2023aloft} or at the feature level (\eg, generate new feature statistics)~\cite{li2021simple}. Traditional image-level augmentation methods usually employ style transfer or adversarial training~\cite{lin2020multi} to generate novel styles. For example, Yue \etal \cite{yue2019domain}  transfer the styles from the images in the ImageNet to the source data, while Zhou \etal \cite{zhou2020deep} employ adversarial training to generate images with unseen styles. However, these image-level augmentation methods often incur computational costs. An alternative perspective is offered by Xu \etal~\cite{xu2020robust}, who observe that the convolution with randomized weights can inherently yield images with novel styles. Different from direct style generation, several methods generate new styles by mixing existing source data. Zhou \etal \cite{zhou2021mixstyle} mix the statistics of two images in the image or feature level while Xu \etal \cite{xu2021fourier} mix the low-frequency components of two images with the Fourier Transformation.

Moreover, various regularization strategies~\cite{zhang2022mvdg,cha2021swad,li2018learning,ding2017deep} have been formulated to mitigate overfitting. The meta-learning training scheme~\cite{li2018learning,li2019episodic,li2019feature} is the most widely employed, which performs regularization with a bi-level optimization process. Li \etal \cite{li2018learning} propose MLDG that splits the source domains into meta-train and meta-test data and then evaluates the model trained on the meta-train data with the meta-test data.  Differently, ensemble learning methods have recently been proposed as an effective approach to achieve better generalization. For example, Cha \etal~\cite{cha2021swad} find that by training a model located in the flat minimum, better generalization ability can be achieved and this can be achieved by simply averaging the weights along the training trajectory. Zhang \etal \cite{zhang2022mvdg} argue that more training trajectories and length can help the regularization and proposes MVDG to exploit the multi-view trajectories. 

Different from previous Domain Generalization methods that employ a sophisticated algorithm to enhance the generalizability, our proposed method is a simple but effective training strategy to identify flat minima with a large learning rate and weight interpolation and can be easily incorporated into previous methods.   

\textbf{Flat minima} have been investigated for a long time~\cite{hochreiter1994simplifying,hochreiter1997flat,kaddour2022fair,zhuang2022surrogate}. It is first introduced in \cite{hochreiter1994simplifying} that the flat minimum can be described with fewer bits than the sharp minimum since the sharp area is more complicated. Thus, according to the principle of Minimum Description Length (MDL), flat minimum generalizes better.  Keskar \etal \cite{keskar2016large} argue that the performance degradation when training with large batch training is due to the sharp minimum it finds while training with a small batch size can find flat minima and generalize better. As flatness is defined according to the weight perturbations, Dinh \etal~\cite{dinh2017sharp} point out that the flatness is sensitive to the reparametrization and Li~\cite{li2018visualizing} propose to apply layer normalization to address this issue. Recently, several attempts have been made to find a flat area during training. SmoothOut~\cite{wen2018smoothout} smoothes the training loss surface by perturbating models with random noise. AMP~\cite{zheng2021regularizing} and SAM~\cite{foret2020sharpness} produce maximum weight perturbations during training to adversarially locate the flat area. Different from explicitly searching for the flat minima, Izmailov \etal \cite{izmailov2018averaging} observe that the weight along the training trajectory always lies on the edge of the loss landscape and the average of them can generate the weights that are located in the flatter area.
Following this observation, recent methods~\cite{cha2021swad,arpit2021ensemble} average the weights along the training trajectory to ensure the model is located in the flat area.
Different from the above methods that introduce weight perturbation~\cite{foret2020sharpness} or dense weight averaging~\cite{cha2021swad}, we provide a novel perspective that incorporates a large learning rate and weight interpolation to effectively identify the flat minima.

\section{Method}

\subsection{Flat Minima Benefit DG}
\begin{figure}[t]
  \hspace{30pt}
  \includegraphics[width=0.65\linewidth]{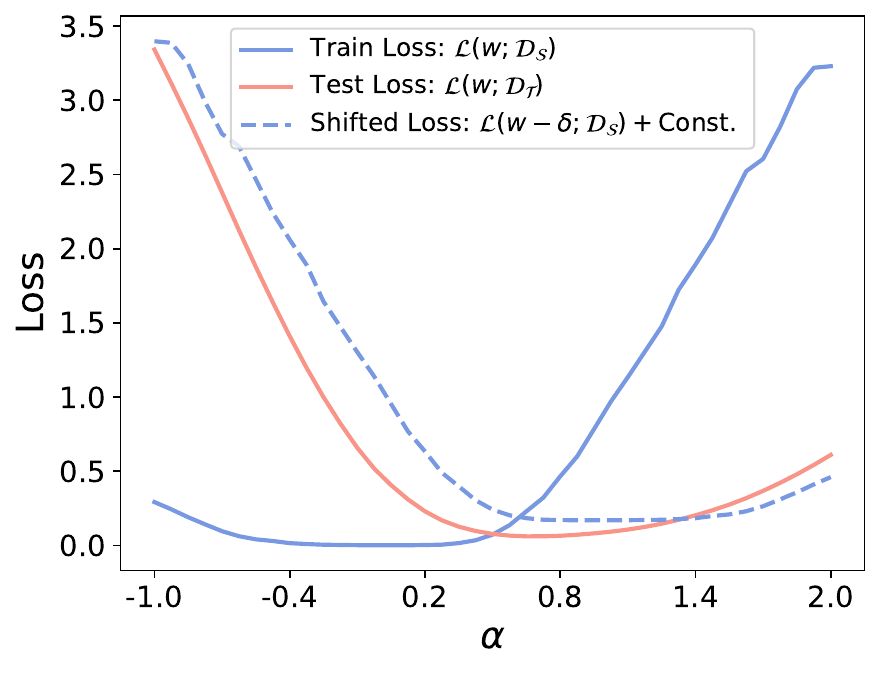}
  \caption{The training, test and shifted training loss landscapes of $w=(1-\alpha) \boldsymbol{\theta}_{\mathcal{S}}^* + \alpha\boldsymbol{\theta}_{\mathcal{T}}^*$ along the direction of $-\boldsymbol{\delta}$.}
  \label{fig_perturbation}
\end{figure}

Let $\mathcal{X}$ and $\mathcal{Y}$ denote the input and label space, respectively. Let $\mathcal{D}_{\mathcal{S}}$ and $\mathcal{D}_{\mathcal{T}}$ be denoted as the training source and test unseen domain without differentiation between individual source domains. The training samples $(x^{tr}, y^{tr})$ are sampled from $\mathcal{D}_{\mathcal{S}}$ with images and their corresponding labels, while the test samples  $(x^{te}, y^{te})$ are sampled from $\mathcal{D}_{\mathcal{T}}$.
A training model is denoted as $f(\cdot|\boldsymbol{\theta}): \mathcal{X} \rightarrow \mathcal{Y}$ parametrized by $\boldsymbol{\theta}$. $\mathcal{L}\left(\boldsymbol{\theta};\mathcal{D}\right)$ is the averaged loss of a domain $\mathcal{D}$. The optimal weights for the source and unseen domain is $\boldsymbol{\theta}_{\mathcal{S}}^*={\textrm{argmin}}_{\boldsymbol{\theta}} \ \mathcal{L}(\boldsymbol{\theta}; \mathcal{D}_{\mathcal{S}})$ and $\boldsymbol{\theta}^{*}_{\mathcal{T}}={\textrm{argmin}}_{\boldsymbol{\theta}} \ \mathcal{L}(\boldsymbol{\theta}; \mathcal{D_{\mathcal{T}}})$.%, respectively.

To validate that searching for flat minima is beneficial to the DG problem, we first introduce the following Theorem, which describes that the perturbations caused by the distributional shift in the input space are equivalent to the perturbations in the weight space, with a constant shift:

\begin{thm}\label{theorem}
  \textit{The distribution shifts of datasets $\mathcal{D}_{\mathcal{S}}$ and $\mathcal{D}_{\mathcal{T}}$ can be equivalently treated as a parameter corruption $\mathbf{v}$ near the corresponding minimum}~\normalfont{\cite{zhang2022ga}}.
  \begin{align}
    \mathcal{L}\left(\boldsymbol{\theta}_{\mathcal{T}}^*+ \mathbf{v}; \mathcal{D}_{\mathcal{T}}\right) \approx \mathcal{L}(\boldsymbol{\theta}_{\mathcal{S}}^*+\mathbf{v} ; \mathcal{D}_{\mathcal{S}})+ \textit{Constant},
  \end{align}
  where $\textit{Constant}=\mathcal{L}\left(\boldsymbol{\theta}_{\mathcal{T}}^*; \mathcal{D}_{\mathcal{T}}\right)-\mathcal{L}(\boldsymbol{\theta}_{\mathcal{S}}^*; \mathcal{D}_{\mathcal{S}})$. Let $\boldsymbol{\delta}=-(\boldsymbol{\theta}_{\mathcal{T}}^*-\boldsymbol{\theta}_{\mathcal{S}}^*)$\textit{, while $\boldsymbol{\theta}$ is near $\boldsymbol{\theta}_{\mathcal{S}}^*$ and $\boldsymbol{\theta}_{\mathcal{T}}^*$. We have}
  \begin{align}
    \mathcal{L}\left(\boldsymbol{\theta}; \mathcal{D}_{\mathcal{T}}\right) \approx \mathcal{L}(\boldsymbol{\theta}+\boldsymbol{\delta} ; \mathcal{D}_{\mathcal{S}})+ \textit{Constant}.
    \label{eq_perturbation}
  \end{align}
\end{thm}
During the test stage, data from the unseen domain are different from the training source domains, which causes the distributional shift. According to \cref{eq_perturbation} in Thm. \ref{theorem}, we know that the distributional shift in the data space (\ie, $\mathcal{D}_{\mathcal{S}}$ and $\mathcal{D}_{\mathcal{T}}$) is equivalent to the perturbations (\ie, $\boldsymbol{\delta}$) in the weight space with a constant shift.
In~\cref{fig_perturbation} on the Art domain of the PACS dataset, we visualize this effect by perturbating the best training weight $\boldsymbol{\theta}_{\mathcal{S}}^*$ to the best test weight $\boldsymbol{\theta}_{\mathcal{T}}^*$, which produces the shifted training loss (dotted line) that is close to the real test loss (red line).
As seen, the training and test data discrepancy results in the loss gap $\mathcal{L}(\boldsymbol{\theta};\mathcal{D}_{\mathcal{T}}) - \mathcal{L}(\boldsymbol{\theta};\mathcal{D}_{\mathcal{S}})$. However, by shifting the parameters $\boldsymbol{\theta}$ along the direction of $\boldsymbol{\delta}$, this gap could be reduced to a constant. Therefore, the distributional discrepancy is approximately equivalent to the parameter perturbations. In other words, the model that is robust to the weight perturbation can resist domain shift. Consequently, if we expect the model trained with source data can generalize well to the unseen domains in DG (\ie, resist the distributional shift during the test), we should ensure that the trained model is located in a flat minimum that is robust to any weight perturbation.
%can resist the weight perturbation, which is equivalent to finding a flat minimum in the weight space where the trained model is insensitive to the weight perturbations caused by the domain gap of unseen test samples.

\subsection{Our Method}

To search for flat minima, different from previous methods that employ ensembling methods (\eg, SWAD~\cite{cha2021swad} or SMA~\cite{arpit2021ensemble}) or adversarial perturbating model weights (\eg, SAM~\cite{foret2020sharpness} or AMP~\cite{zheng2021regularizing}), we employ fast-slow weight updating strategy, namely Lookahead, combined with a large learning rate for the training. Assume that we are training a weight $\boldsymbol{\theta}_i$ in the $i$-th iteration with a learning rate of $\eta$. Before each iteration starts, we initialize the fast weight with $\boldsymbol{\theta}_i^1=\boldsymbol{\theta}_i$ and then train it with a large learning rate $\eta$ for $k$ steps and obtain $\boldsymbol{\theta}_i^k$.
Then the new weight $\boldsymbol{\theta}_{i+1}$ in the next iteration can be obtained by interpolating the fast $\boldsymbol{\theta}_{i}^k$ and slow weight $\boldsymbol{\theta}_i$:
\begin{align}
  \boldsymbol{\theta}_{i+1}=(1-\alpha)\boldsymbol{\theta}_{i}+\alpha\boldsymbol{\theta}_{i}^{k},
  \label{eq_meta_learning}
\end{align}
where {$\alpha$ is the interpolation ratio}.
This training trajectory works similarly to the meta-learning algorithm~\cite{finn2017model} that trains the task-specific weight in the inner loop and then updates the original model in the outer loop with the trained weight. Therefore, in the following sections, we call the training process of the fast/slow weight as the training in the inner/outer loop. Benefiting from the large learning rate and weight interpolation that help produce diverse weights and identify flat minima, this simple and effective algorithm can achieve surprisingly good generalization ability without incurring extra computational costs compared to ERM. Therefore, in the following sections, we will detail the effectiveness of the large learning rate and weight interpolation in finding flat minima and producing diverse weights.

\subsection{Large Learning Rate Promotes Diversity}

To achieve satisfactory performance, the diversity~\cite{rame2022diverse} of weights plays a critical role in methods employing the weight averaging technique~\cite{cha2021swad,arpit2021ensemble}. However, utilizing a small learning rate cannot provide sufficiently large diversity compared to a large learning rate. To confirm this, we conducted an experiment that starting from the same initial weight $\boldsymbol{\theta}$, we train two weights using a large learning rate, \ie, $5e-4$ and a commonly employed~\cite{gulrajani2020search} small learning rate, \ie, $5e-5$, for $k$ ($k=15$) steps with the same training data, and obtains $\boldsymbol{\theta}_{\text{large}}^k$ and $\boldsymbol{\theta}_{\text{small}}^k$, respectively. Then we compare the feature diversity~\cite{kornblith2019similarity} and prediction diversity~\cite{aksela2003comparison} between the two trained weights and the initial weight $\boldsymbol{\theta}$.
The feature diversity is obtained by comparing the representation from the penultimate layer before the classifier layer with the Centered Kernel Alignment~(CKA) method. The prediction diversity is measured with $N_{\text{diff}}/N_{\text{simul}}$, where $N_{\text{diff}}$ is the number of different predictions and $N_{\text{simul}}$ is the number of same prediction. As shown in \cref{tab_diversity_compare}, by training with a large learning rate, the diversity of the features and predictions is larger than the small learning rate, which indicates that the large learning rate can effectively promote the diversity of the trained weights. Besides the diversified weights, a large learning rate can also help identify flat minima, which will be discussed in the following section.

\begin{table}[t]
  \begin{center}
    \caption{Feature and prediction diversity comparison between two weights differed in the training learning rate.}
    \resizebox{1\columnwidth}{!}{
      \begin{tabular}{ l |  c c c c}
        \toprule
        \multirow{2}{*}{Diversity} & \multicolumn{2}{c}{Features$\downarrow$} & \multicolumn{2}{c}{Predicitons$\uparrow$}                          \\
                                   & in-domain                                & out-domain                                & in-domain & out-domain \\
        \midrule
        Small LR (5e-5)            & 0.9822                                   & 0.9752                                    & 0.0009    & 0.007      \\
        Large LR (5e-4)            & 0.7882                                   & 0.5859                                    & 0.0024    & 0.028      \\
        \bottomrule
      \end{tabular}
    }
    \label{tab_diversity_compare}
  \end{center}
  \vspace{-5pt}
\end{table}

\subsection{Large Learning Rate Favors Flat Minima}
\label{sec_lr}

% To find flat minimum, previous methods either employ explicit methods that add adversarial weight perturbations to the model during training or adopt a weight averaging strategy to obtain ensembled weights. In this paper, 
Different from previous methods using a sophisticated algorithm to find flat minima, adjusting the learning rate not only helps produce diverse weights but also helps identify flat minima~\cite{seong2018towards,mohtashami2022avoiding,smith2019super}. To analyze this phenomenon, we assume a quadratic model~\cite{zhang2019lookahead} for simplicity:
\begin{align}
  \mathcal{L}^{q}(\boldsymbol{\theta})=\frac{1}{2}(\boldsymbol{\theta}-\mathbf{c})^T \mathbf{H}(\boldsymbol{\theta}-\mathbf{c}),
\end{align} where $\mathbf{c} \sim \mathcal{N}\left(\boldsymbol{\theta}^*, \Sigma\right)$, $\mathbf{H}$ and $\Sigma$ are diagonal and that, without loss of generality, $\boldsymbol{\theta}^* = 0$. Assume that $h_{max}$ is the maximum eigenvalue of $\mathbf{H}$, which indicates the flatness of the current model (the smaller, the flatter). The updating rule of $\boldsymbol{\theta}$ is $\boldsymbol{\theta}_{t+1}=\boldsymbol{\theta}_t-\eta \nabla {\mathcal{L}^q}(\boldsymbol{\theta} ; \mathbf{x})$, where $t$ is iteration number and $\eta$ is learning rate. Then, the convergence of the training algorithm criteria requires:
\begin{align}
  0 < h_{\max} <\frac{2}{\eta}.
\end{align}
% \mathcal{L}^{q}(\boldsymbol{\theta}+\epsilon) - \mathcal{L}^{q}(\boldsymbol{\theta})  \approx 
{Although the neural network training objectives are not globally quadratic, the second-order Taylor approximation around any point $\boldsymbol{\theta}$ in parameter space is approximately a quadratic function whose ``$\mathbf{H}$'' matrix is the Hessian at $\boldsymbol{\theta}$~\cite{cohen2021gradient}.}  Consequently, by increasing the learning rate $\eta$, to enable the stability of the training process,  the model will reach $\boldsymbol{\theta}$ where the maximum eigenvalue $h_{\mathrm{max}}$ of second-order gradients is smaller, which indicates the training model is at a flatter area. Intuitively, with a large learning rate, if $h_{\mathrm{max}}$  is large (\ie, the loss surface is sharp), it is easier for the training model to escape from this local sharp minimum and jump to a flatter area, as illustrated in \cref{fig_intro}.

\begin{figure}[t]
  \subfloat{
    \includegraphics[width=0.48\linewidth]{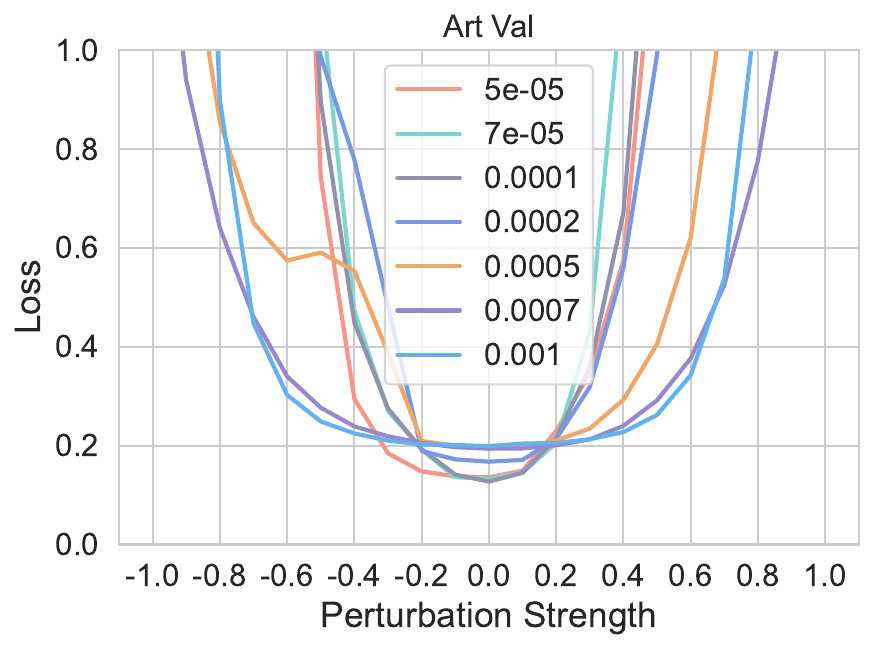}
  }
  \subfloat{
    \includegraphics[width=0.48\linewidth]{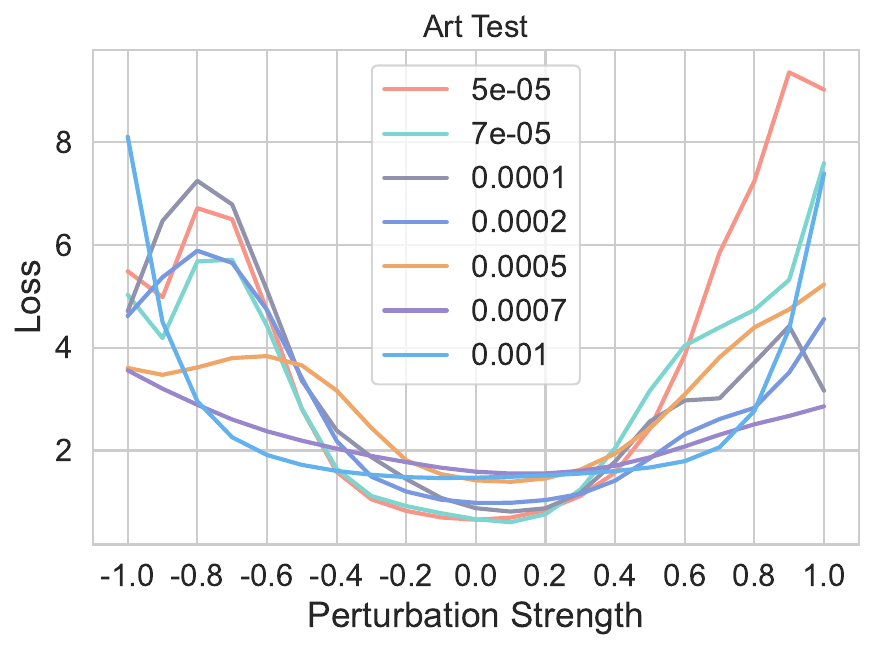}
  }
  \caption{Flatness changes according to the learning rate. With a larger learning rate, the model can find a flatter area while the performance on the unseen domain degrades.}
  \label{fig_lr}
  % \vspace{-5pt}
\end{figure}

We plot the loss surfaces of several models trained with the same configurations that differ only in the learning rate on the PACS dataset in \cref{fig_lr}. As shown from the left figure, with the learning rate increases, the loss curvatures become flatter, which means the corresponding model is located in a flatter area that generalizes better. However, although the model trained with a larger learning rate favors the flat minimum, the validation loss (\ie, `0.0' in the left figure of \cref{fig_lr}) is also larger than the models trained with smaller learning rates (\eg, the large loss with a learning rate of $0.001$),  which means simply training with a larger learning rate hurts the ability to fit the source domain. An intuitive example is a constant function that is completely robust to perturbations but has no classification ability.  Therefore, classification and generalization abilities should be optimized together.

While weight averaging can enhance the model's robustness, it does not provide a guarantee of model convergence. In order to take advantage of both a large learning rate, which generates diverse weights and uncovers flat minima, and the benefits of weight averaging, we introduce the Lookahead training strategy. Lookahead shares a similar concept with weight averaging but achieves a favorable balance between optimization and generalization~\cite{zhou2021towards}. This is accomplished through the interpolation of fast and slow weights using a small coefficient to ensure that the updated slow weight is not adversely affected by the unconverged fast weight.
In this way, the relationship between the two learning rates (\ie, $\eta$ and $\alpha$) and the eigenvalue $h_{\max}$ has been changed as follows (details are in Appendix):
\begin{align}
   h_{\max} < \frac{1}{\eta}(\frac{1}{\alpha})^{1 / k}+\frac{1}{\eta} 
\end{align} where $k$ is the length of the inner loop to train the fast weight.
As seen, when $\eta$ is fixed, $\alpha=1$ means a normal training process, which suffers an optimization problem during training. When decreasing $\alpha$, the problem is reduced while the training model is easily stuck in a sharp minimum since the right-hand formulation becomes larger. Therefore, introducing an additional interpolation ratio $\alpha$ can inevitably increase $h_\max$ and sacrifice the flatness found by the algorithm. However, by adopting a large length $k$ (\eg, $k=15$) of the inner loop, this effect can be reduced, which ensures that this algorithm can benefit both from the large learning rate for better generalization and the interpolation training paradigm for better optimization.

\label{sec_longer}

\subsection{Weight Interpolation Identifies Flat Minima}

\begin{figure}[t]
  \subfloat{
    \includegraphics[width=0.49\linewidth]{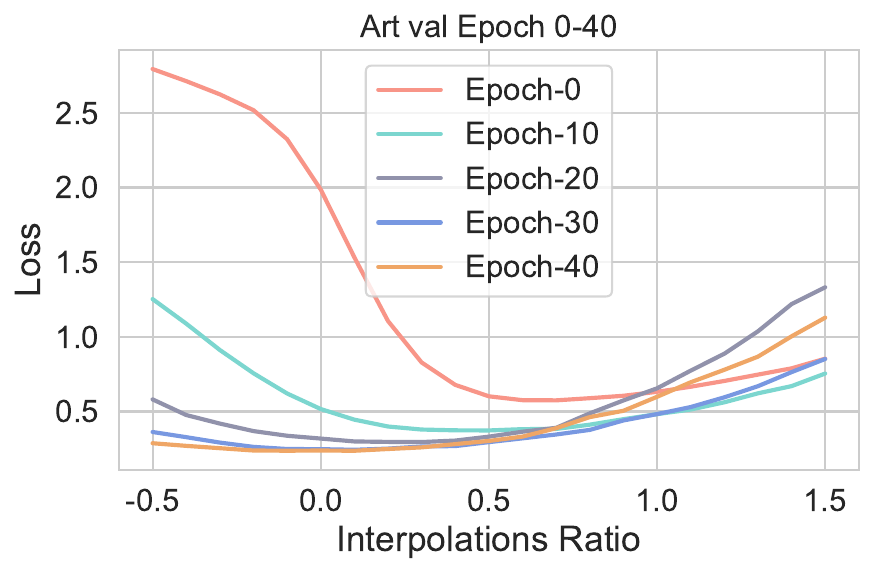}
  }
  \subfloat{
    \includegraphics[width=0.49\linewidth]{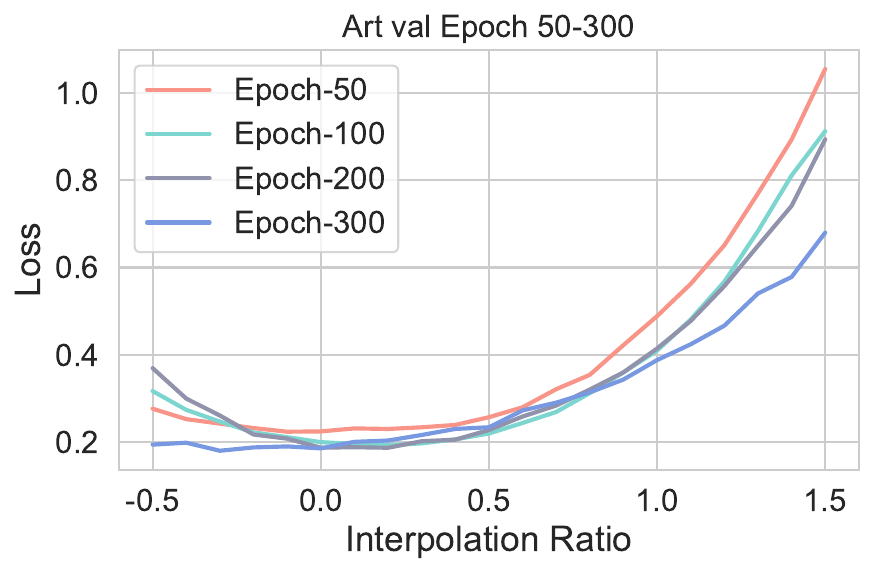}
  }
  \caption{The loss surface of the interpolated model weights ($\boldsymbol{\theta}=(1-\alpha) \boldsymbol{\theta}_0 + \alpha\boldsymbol{\theta}_1$) from different training epochs on the Art domain. The start and end weights of fast weight are at `0.0' and `1.0', respectively.}
  \label{fig_interpolation}
  % \vspace{-5pt}
\end{figure}

Besides a large learning rate, interpolation not only helps ease optimization but also plays a critical role in finding flat minima. It has been observed in the weight ensemble learning methods (\eg, SWA~\cite{izmailov2018averaging}, SWAD~\cite{cha2021swad}) that interpolated weights from two model weights that share a partial training trajectory tend to remain in the flat area of the loss landscape. To investigate the role of interpolation in Lookahead, we interpolate the start weight (same as the slow weight from the previous iteration) and end weight of the fast weight at different epochs. As shown in \cref{fig_interpolation}, at the beginning of training, the start weight has a high loss while the end weight is at a flatter area with a relatively low loss. Interpolation between these two weights leads to a decrease in loss and a flatter area. As training progresses, the start weight achieves low loss in a flatter area, while the loss of the end weight remains high. It is because the end weight is trained with a large learning rate starting from the first weight, which prevents the end weight from converging with a low loss. Therefore, directly taking the end weight as the final training weight (\ie, slow weight) causes an optimization problem, as discussed in \cref{sec_lr}. However, by choosing a relatively small interpolation ratio (\eg, $0.05$), the interpolation between the start and end weight can ease the optimization problem and help identify flat minima.

\label{sec_entropy}
An intuitive explanation is that since the end weight is located in a flat area, with the interpolation, the start weight is pulled towards these flat areas instead of stuck in a local sharp minimum. Theoretically, Lookahead with weight interpolation optimizes the local entropy loss~\cite{chaudhari2019entropy}, which measures the landscape flatness of the current weight. It is defined as:
\begin{align}
  F(\boldsymbol{\theta}, \gamma)=\log \int_{\boldsymbol{\theta}^{\prime}} \exp \left(-\mathcal{L}\left(\boldsymbol{\theta}^{\prime}\right)-\frac{\gamma}{2}\left\|\boldsymbol{\theta}-\boldsymbol{\theta}^{\prime}\right\|_2^2\right) d \boldsymbol{\theta}^{\prime},
\end{align}
where $\boldsymbol{\theta}'$ is the neighborhood of current weight and $\gamma$ defines its radius. $\mathcal{L}$ is a loss function defined on any given dataset. By maximizing this objective function (\ie, minimizing $-F(\boldsymbol{\theta}, \gamma)$), the derivative is as follows :
\begin{align}
  \nabla_x F\left(\boldsymbol{\theta}, \gamma \right)                                                                & =\gamma\left(\boldsymbol{\theta}- \mathbb{E}[\boldsymbol{\theta}^{\prime}] \right), \\
  \label{eq_derivative}
  \boldsymbol{\theta}_{i+1} = \boldsymbol{\theta}_i - \gamma(\boldsymbol{\theta}_i-\mathbb{E}[\boldsymbol{\theta}']) & =(1-\gamma)\boldsymbol{\theta}_i+\gamma \mathbb{E}[\boldsymbol{\theta}'].
\end{align}
Therefore, the updating rule of maximizing this objective function in \cref{eq_derivative} takes approximately the same form as the interpolation in the training scheme of Lookahead in \cref{eq_meta_learning}, except that the expectation is replaced with an empirical estimation. The average operation can reduce the noise of short trajectories and produce an accurate gradient estimation.
Instead, since the expectation of the neighborhood weights of current weight is required which cannot be obtained directly due to the large search space in the weight space,  Entropy-SGD~\cite{chaudhari2019entropy} employs the SGLD algorithm with a Markov chain Monte-Carlo (MCMC) technique to efficiently sample from the neighborhood weights of the current weight. This sampling technique pulls the training weight not too far away from the current weight and adds noise during sampling.
However, since the minibatch in the SGD naturally contains the noise~\cite{palacci2018scalable}, we find that this additional noise does not improve the performance in the experiments and brings extra hyperparameter burden.
In contrast, Mandt \etal \cite{mandt2017stochastic} have shown that stochastic gradient descent (SGD) itself can be viewed as a sampling method to approximate the target distribution.
As a result, by simply adopting SGD for fast weight training, we can approximate the expectation in \cref{eq_derivative} and maximize the local entropy.
Besides, by training the fast weight for more steps, the estimation can be more accurate with better performance.

\label{sec_noise}

% \section{Method}

\begin{figure}[t]
  \centering
  \includegraphics[width=0.85\linewidth]{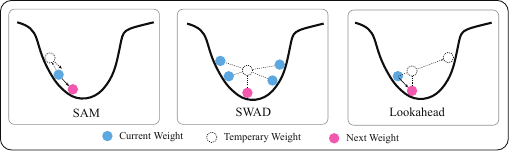}
  \caption{Illustration of different flat minima searching methods (\ie, SAM, SWAD, and Lookahead) on a loss landscape.}
  \label{fig_SAM_compare}
  % \vspace{-5pt} 
\end{figure}

\subsection{Comparison to SAM and SWAD}

In the realm of discovering flat minima, two conventional methods, namely SWAD~\cite{cha2021swad} and SAM~\cite{foret2020sharpness}, exist. However, our proposed mehtod provides a new perspective on the large learning rate to uncover flat minima, which differs significantly from these methods.
The essence of SWA's flat minima discovery lies in its dense ensemble of model weights, enabling an effective shift towards a shared area, which minimizes the distance to all the training weights. Consequently, it exhibits robustness against perturbations.
On the other hand, SAM takes a distinctive route by explicitly ascending from the original weight to identify a local maximum and then `pushing' the weight away from this point by applying the gradient of the local maximum to the original weight.
Contrastingly, Lookahead employs an entirely different strategy to locate flat minima, as depicted in \cref{fig_SAM_compare}. It initiates a search for flat minima using a larger learning rate and subsequently `pulls' the training weight towards this identified region. As a result, Lookahead offers a novel perspective on the exploration of flat minima. Furthermore, Lookahead gains an advantage from its utilization of a larger learning rate, enabling it to explore a broader region of the weight space. In contrast, both SWAD and SAM are limited to exploring smaller regions, which could potentially lead to sub-optimal solutions. In addition, SAM's training scheme pushes the model away from the initial point, potentially leading to knowledge forgetting of the original pre-trained model. Differently, SWAD and Lookahead avoid this issue by interpolating with the training weights, leading to better performance for DG.

Besides, in contrast to Reptile~\cite{nichol2018reptile}, this work diverges in two key aspects. Firstly, Reptile optimizes the same data within a training trajectory, a strategy prone to overfitting in the domain generalization (DG) setting. In contrast, we employ a different batch for optimization in each iteration within a trajectory, enhancing the model's ability to generalize across domains. Secondly, Reptile employs a small learning rate, following a conventional training scheme. However, this approach does not effectively aid in discovering flat minima and achieving superior generalization, which will be validated in \cref{sec_exp}.

\begin{figure}[t]
  \centering
  \includegraphics[width=1.\linewidth]{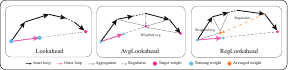}
  \caption{Method comparison between our proposed Lookahead, AvgLookahead, and RegLookahead.}
  \label{fig_framework} 
  % \vspace{-5pt}
\end{figure}

\subsection{Regularize Lookahead}

% However, simply applying Lookahead to train the model suffers from continuous degradation when it has already been on the flat area. As mentioned above, as training goes on, the model tends to stay in the flat area. However, in the latter stage, the performance of the model would fluctuate and degrade due to the large learning rate employed in the inner loop. Although the large learning rate could favor the flat area, Besides, only utilizing the end weight also causes performance fluctuation since it may not be the optimal weight for interpolation. 
% First,  employing the end weight for the interpolation may not be optimal and would cause the performance to fluctuate since a single weight is noisy. Besides, as the training goes on, the algorithm also has a tendency to overfit the training data, which also causes a poor generalization ability and poor correlation of training and testing accuracy.  
Although the simple Lookahead algorithm can achieve good generalization performance by finding a flat minimum, it is still easy to overfit the training data in the later stage, where the training loss decreases to a small value and the corresponding gradients are also small. To prevent this problem, we propose two different strategies namely AvgLookahead and RegLookahead as shown in \cref{fig_framework}.

\textbf{AvgLookahead}: There are two drawbacks of interpolation only with the end weights: 1) instability of training, and 2) quickly overfitting the training data by moving far away from the initial weight. Therefore, instead of only employing the end weight for interpolation, we propose to perform the weighted average along the weights in the inner loop of Lookahead to both stabilize training and prevent overfitting:
\vspace{-5pt}
\begin{align}
  \boldsymbol{\theta}_{i+1} = (1-\alpha) * \boldsymbol{\theta}_{i} + \alpha * \sum_{j}^k \beta_{j}\boldsymbol{\theta}_i^j, \ \ \ \ \  \sum_j^k\beta_j=1.
\end{align}
AvgLookahead can achieve a lower variance with the following guarantee (details are in Supplementary Material):
\begin{align}
  V^*_{\text{AvgLA}} & = \frac{\alpha^2 Y}{\mathbf{I} - \left[(1-\alpha)\mathbf{I}+\alpha  \sum_{i=0}^{k-1}  \beta_i\mathbf{M}^i\right]^2 } V_{ERM}^* \\
                     & \le V^*_{\text{LA}} \le V^*_{\text{ERM}},
\end{align}
where $\mathbf{M} = (\mathbf{I}-(\mathbf{I}-\eta \mathbf{H}))$, $Y=\sum_{i=0}^{k-1}  \beta_i^2 \left(\mathbf{I}-\mathbf{I}^{2i}\right)+2  \sum_{i=0}^{k-1} \sum_{j=0}^{i-1} \beta_i \beta_j \mathbf{I}^{i-j} (\mathbf{I}-\mathbf{M}^{2j})$. Note that we simply set $\beta_i=\frac{1}{k}$ to reduce the hyperparameter, which is found to work well across different datasets and backbones.

\textbf{RegLookahead}: Besides the weighted average in the inner loop, introducing an explicit regularization operation during training can also help prevent the overfitting problem. We employ the averaged weights along all previous weights $\boldsymbol{\theta}_i^{\textrm{avg}}=\frac{1}{l}\sum_{m=1}^l\boldsymbol{\theta}_i^{m}$ as the target to regularize the training weights with L2 norm in the inner loop as follows:
\begin{align}
  \boldsymbol{\theta}_{i}^{j+1} = \boldsymbol{\theta}_{i}^{j} - \eta \nabla (\mathcal{L}(\boldsymbol{\theta}_i^j) + \lambda ||\boldsymbol{\theta}_{i} - \boldsymbol{\theta}_i^{\textrm{avg}}||_2^2),
\end{align}
where $\lambda$ is the regularization strength. By employing these two different regularizations, the trained model can obtain better generalizability with little computational overhead.
% Besides, directly applying the regularization in the whole training stage could result in the underfitting problem, inspired by~\cite{cha2021swad}, we employ the regularization when the model starts to overfit the data by detecting its validation loss changes. 
% Note that, simply employing AvgLookahead or RegLookahead could inevitably degrade performance when the model does not hurt the performance. Therefore, during optimization, we add noise into the model to further encourage the exploration and prevent the underfitting problem caused by AvgLookahead or RegLookahead. Inspired by \cite{wen2018smoothout}, we employ the filter-level noise: $\hat{\epsilon^i} = s\frac{\|w^i\|}{\|\epsilon^i\|} \epsilon^i$, where $\epsilon\in\mathcal{U}(-s, s)$ and $w^i$ is a filter in the model.

% However, when applying Lookahead to traditional classification tasks, although it can find a flat minimum than the trained model, it cannot find a flatter area than SAM. We hypothesize that the learning rate scheduling strategy during training plays a critical role. At the early stage, with a large learning rate, Lookahead can quickly achieve better performance and flat area than SGD and SAM. However, when the learning rate decays, the benefit of a large learning rate diminishes.

\section{Experiments}

\begin{table}[t]
  \begin{center}
    \caption{Algorithm-specific hyperparameter search space.}
    \resizebox{1\columnwidth}{!}{
      \begin{tabular}{ l |  c c c c | c}
        \toprule
        {Parameter}                       & {Default value} & {Searched grids}                      \\
        \midrule
        learning rate $\eta$              & 5e-4            & [5e-4, 1e-3, 5e-3]                    \\
        interpolation ratio  $\alpha$     & 0.05            & [0.01, 0.03, 0.05]                    \\
        length of inner loop $k$          & 15              & [5, 10, 15]                           \\
        \hline
        regularization strength $\lambda$ & 0.01            & [0.001, 0.005, 0.01, 0.03, 0.05]    \\
        number of averaging weight        & 10              & [1, 5,10]                             \\
        \bottomrule
      \end{tabular}
    }
    \label{tab_hyper}
  \end{center}
  \vspace{-5pt} 
\end{table}

\begin{table*}[h]
  \begin{center}
    \caption{Comparison to SOTA with Resnet-50 and ResNeXt-50 as backbones with \textit{accuracy} ($\%$) and training \textit{time} (min.). RegLA and AvgLA are shorthands of RegLookahead and AvgLookahead while En. means ensemble three different runs. The best performance is marked as \textbf{bold}. Methods marked with $^\dagger$ are reproduced results.}
    \resizebox{2.\columnwidth}{!}{
      \begin{tabular}{ l | c |  c| c  c c c c | c}
        \toprule
        Algorithm                           & Venue      & Time    & PACS                  & VLCS                  & OfficeHome            & TerraIncognita        & DomainNet             & Avg.            \\
        \bottomrule
        \multicolumn{9}{c}{\textbf{Resnet-50}~\cite{he2016deep}}                                                                                                                                             \\
        \toprule
        ERM                                 & -          & $32.1$  & $85.6$                & $77.5$                & $66.5$                & $46.1$                & $40.9$                & $63.3$          \\
        ERM$^\dagger$                       & -          & $32.1$  & $83.0\pm0.2$          & $77.7\pm0.3$          & $68.1\pm0.1$          & $47.8\pm0.7$          & $44.0\pm0.1$          & $64.2$          \\
        IRM~\cite{arjovsky2019invariant}    & ICML-19    & -       & $83.5$                & $78.6$                & $64.3$                & $47.6$                & $33.9$                & $61.6$          \\
        MLDG~\cite{li2018learning}          & AAAI-18    & -       & $84.9$                & $77.2$                & $66.8$                & $47.8$                & $41.2$                & $63.6$          \\
        SAM~\cite{foret2020sharpness}       & ICLR-22    & $55.6$  & $85.8$                & $79.4$                & $69.6$                & $43.3$                & $44.3$                & $64.5$          \\
        SWAD~\cite{cha2021swad}             & NeurIPS-21 & $32.1$  & $88.1$                & $79.1$                & $70.6$                & $50.0$                & $46.5$                & $66.9$          \\
        MVDG$^\dagger$~\cite{zhang2022mvdg} & ECCV-22    & $90.9$  & $88.3$                & $78.8$                & $71.2$                & $53.7$                & $45.5$                & $67.5$          \\
        EoA~\cite{arpit2021ensemble}        & NeurIPS-22 & $96.3$  & $88.6$                & $79.1$                & ${72.5}$              & $52.3$                & ${47.4}$              & $68.0$          \\
        \rowcolor{Gray} Lookahead           & Ours       & $34.9$  & $88.1\pm0.3$          & $79.1\pm0.1$          & $71.3\pm0.2$          & $51.2\pm0.1$          & $45.9\pm0.1$          & $67.1$          \\
        \rowcolor{Gray} RegLookahead        & Ours       & $34.6$  & $88.3\pm0.7$          & $79.3\pm0.5$          & $71.4\pm0.2$          & ${53.6\pm0.7}$        & $45.8\pm0.1$          & $67.7$          \\
        \rowcolor{Gray} AvgLookahead        & Ours       & $34.9$  & ${88.8\pm0.4}$        & ${79.6\pm0.5}$        & $71.6\pm0.2$          & $53.0\pm0.0$          & $45.7\pm0.1$          & $67.7$          \\
        \rowcolor{Gray} RegLA + En.         & Ours       & $104.7$  & $\mathbf{89.3}$       & $79.8$                & $72.7$                & $\textbf{54.2}$       & $\textbf{47.32}$      & $\textbf{68.7}$ \\
        \rowcolor{Gray} AvgLA + En.         & Ours       & $104.7$  & $89.1$                & $\textbf{79.9}$       & $\textbf{73.0}$       & $53.8$                & $47.1$                & $68.6$          \\
        \bottomrule
        % \multicolumn{8}{c}{{\textbf{GFNet}}}                                                                                                    \\
        % \bottomrule
        % ERM^$\dagger$                       &            & $88.7\pm0.5$ & $79.1\pm0.5$ & $72.2\pm0.1$ & $50.0\pm1.2$ & $47.22\pm0.2$  & $67.44$ \\
        % SWAD^$\dagger$                      & NeurIPS-21 & $89.5\pm0.2$ & $80.1\pm0.1$ & $72.7\pm0.2$ & $55.3\pm0.4$ & $47.57\pm0.0$  & $69.03$ \\
        % \bottomrule
        % \rowcolor{Gray} Lookahead           & Ours       & $91.1\pm0.1$ & $80.0\pm0.3$ & $75.2\pm0.1$ & $54.8\pm0.1$ & $50.89 \pm0.1$ & $70.40$ \\
        % \rowcolor{Gray} RegLookahead        & Ours       &              &              &              &              &                &         \\
        % \rowcolor{Gray} AvgLookahead        & Ours       &              &              &              &              &                &         \\
        % \toprule
        \multicolumn{9}{c}{{\textbf{ResNeXt-50}~\cite{yalniz2019billion}}}                                                                                                                                   \\
        \toprule
        ERM$^\dagger$                       &            & $41.3$  & $90.8\pm0.5$          & $79.6\pm0.3$          & $75.8\pm0.2$          & $47.6\pm0.9$          & $49.3\pm0.1$          & $68.6$          \\
        SWAD$^\dagger$~\cite{cha2021swad}   & NeurIPS-21 & $41.3$  & $94.0\pm0.1$          & $80.6\pm0.1$          & $79.2\pm0.0$          & $53.0\pm0.3$          & $48.7\pm0.1$          & $71.1$          \\
        EoA~\cite{arpit2021ensemble}        & NeurIPS-22 & $123.9$ & $93.2$                & $80.4$                & $80.2$                & $55.2$                & $54.6$                & $72.7$          \\
        \rowcolor{Gray} Lookahead           & Ours       & $42.5$  & $93.6\pm0.1$          & $80.5\pm0.2$          & $79.7\pm0.1$          & $57.0\pm0.2$          & $51.7\pm0.0$          & $72.5$          \\
        \rowcolor{Gray} RegLookahead        & Ours       & $42.5$  & $\mathbf{94.2\pm0.2}$ & $\mathbf{81.9\pm0.1}$ & $80.2\pm0.1$          & $\mathbf{57.4\pm0.9}$ & $52.4\pm0.2$          & $73.2$          \\
        \rowcolor{Gray} AvgLookahead        & Ours       & $42.5$  & $94.1\pm0.3$          & $81.6\pm0.1$          & $\mathbf{80.3\pm0.2}$ & $57.2\pm0.1$          & $\mathbf{53.6\pm0.0}$ & $\mathbf{73.4}$ \\
        % \toprule
        % \bottomrule
        % \rowcolor{Gray} RegLA + En.         & Ours       & 32   &                                                                                                                                         \\
        % \rowcolor{Gray} AvgLA + En.         & Ours       & 32   &                                                                                                                                         \\
        \toprule
      \end{tabular}
      \label{tab_sota}
    }
  \end{center}
  \vspace{-15pt}
\end{table*}

In the following experiments, we first provide the details of the experiments. Subsequently, we engage in a comprehensive comparison of our proposed methodologies against the current state-of-the-art techniques on both \textit{classification} and \textit{semantic segmentation} tasks. Following this, we conduct an ablation study on our proposed method to validate the effectiveness of the large learning rate, interpolation, and two regularizations. We then proceed to a comparative evaluation, pitting Lookahead against SAM and SWAD, to illuminate the respective performances of these methods in the context of discovering flat minima. Finally, we provide further analysis for a better understanding of our methods.

\textbf{DataSets.} In line with the approach of Gulrajani and Lopez-Paz~\cite{gulrajani2020search}, we assess the performance of our method using the DomainBed benchmark for the \textit{classification task}, encompassing a diverse array of datasets: \textbf{PACS}~\cite{li2017deeper} with 9,991 images, 7 classes and 4 domains, \textbf{VLCS}~\cite{torralba2011unbiased} with 10,729 images with 5 classes and 4 domains, \textbf{OfficeHome}~\cite{venkateswara2017deep} with 15,500 images with 65 classes, \textbf{TerraIncognita}~\cite{beery2018recognition} with 24,788 images, 10 classes and 4 domains, and \textbf{DomainNet}~\cite{peng2019moment} with 586.575 images with 126 classes and 6 domains. There is a large domain gap between domains of PACS, TerraIncognita, and DomainNet datasets, while for VLCS and OfficeHome, it is smaller. For the \textit{semantic segmentation} task, the model is trained on a synthetic \textbf{GTAV}~\cite{richter2016playing} (G) dataset that contains 12403, 6382, and 6181 images for training, validation, and test sets, respectively. And then adapted to the other three real-world datasets, including \textbf{Cityscapes}~\cite{cordts2016cityscapes} (C), \textbf{BDD100K}~\cite{yu2020bdd100k} (B) and \textbf{Mappilary}~\cite{neuhold2017mapillary} (M). They consist of 2975, 7000, and 18000 images for the training set and 500, 1000, and 2000 for the validation set.

\textbf{Implementation details.} For the \textit{classification} task, following DomainBed benchmark~\cite{gulrajani2020search}, we employ ImageNet~\cite{deng2009imagenet} pre-trained Resnet-50~\cite{he2016deep} as our default backbone if not especially mentioned. The inner loop of Lookahead is optimized with the Adam optimizer, while the outer optimizer is vanilla SGD without momentum and weight decay (\ie, simple weight interpolation). The batch size is $126$ where each domain contains $32$ images. Note that, since our method employs an inner loop in each iteration during training, for a fair comparison, we divide the original training steps by the length of the inner loop for a fair comparison. Thus, no additional computational cost is incurred. We use the random searched default parameters of dropout and weight decay from SWAD~\cite{cha2021swad}. We adopt the default augmentation for training that comprises the random resized crop with scale factor within $[0.7, 1.0]$, random horizontal flip with a probability of $0.5$, color jittering with a strength of $0.3$, and random grayscale with a probability of $0.1$.  For the \textit{semantic segmentation} task, we adopt ImageNet~\cite{deng2009imagenet} pre-trained Resnet-50~\cite{he2016deep} as our Backbone and DeepLabV3+~\cite{chen2018encoder} as the segmentation network. Following previous settings~\cite{choi2021robustnet}, the optimizer is SGD with a momentum of $0.9$ and weight decay of $5$e$-4$ except that its learning rate of $\eta$ is $0.05$ (instead of commonly used $0.01$) while we set $\alpha=0.05$. The learning rate is decreased using the polynomial policy with a power of $0.9$ The batch size is $8$. The augmentation is random scaling within a range of $[0.5, 2.0]$ and random cropping with a size of $768\times 768$. The model is trained in $40k$ iterations.

\textbf{Evaluation protocol.} For the \textit{classification} task, we designate one domain as the unseen domain, while treating the remaining domains as source domains. For a fair comparison, we employ the same hyperparameter selection as previous methods that use a validation set of source domains to select the best hyperparameters for our method. Note that for efficiency, we adopt the best hyperparameter of ERM in SWAD~\cite{cha2021swad} and then search for the rest hyperparameter of Lookahead (\ie, the learning rate $\eta$, the interpolation ratio $\alpha$ and the step length $k$ of inner loops). Similarly, for the AvgLookahead and RegLookahead, we also search based on the hyperparameters searched from Lookahead and search for the rest of hyperparameters (\ie, regularization strength $\lambda$ and the number of averaging weights). The search space is listed in \cref{tab_hyper} All reported experiments are repeated three times and averaged. Other details (\eg, network architecture, augmentation, hyperparameters selection). The best model is selected with the validation set. For the \textit{semantic segmentation} task, we train the model on the synthetic GTAV dataset and test it on the other three datasets (\ie, Cityscapes, BDD100K, and Mappilary). The best hyperparameter is selected with the validation set and we use the last model for evaluation.

\subsection{Comparison to State-Of-The-Art Methods}

\subsubsection{Classification task}

We compare the accuracy and time cost of our method to several typical state-of-the-art methods on the DomainBed benchmark~\cite{gulrajani2020search}. Specifically, IRM~\cite{arjovsky2019invariant} learns domain-invariant feature, MLDG~\cite{li2018learning} and MVDG~\cite{zhang2022mvdg} both employ the meta-learning training scheme, SWAD~\cite{cha2021swad} and EoA~\cite{arpit2021ensemble} both ensemble training models, and SAM~\cite{foret2020sharpness} aims to  find flat minima. It's important to note that the listed time corresponds to the duration required to complete the training of a single model.
Our results, as summarized in \cref{tab_sota}, demonstrate that Lookahead can significantly surpass the ERM baseline and MLDG, the first meta-learning algorithm in DG, which validates its effectiveness. Besides, It also can achieve better performance than SWAD and SAM. In terms of SAM, it cannot even defeat ERM on PACS and TerraIncognita datasets. The backbone (\ie, Resnet-50) is pre-trained on ImageNet~\cite{deng2009imagenet}, which already contains diverse useful information for generalization. The high performance of ERM relies heavily on the pre-trained model. However, since the training process of SAM encourages the weight escape from the local minimum, the final trained weight tends to leave far away from the original weight and thus, performs poorly.  Different from SAM which only searches for the flat area, Lookahead also tries to not leave far away from the pre-trained model, resulting in better performance ($67.1\%$ vs. $64.5\%$) on these two datasets.

Besides, Lookahead demonstrates performance that is on par with MVDG and EoA. It's worth noting that these two methods ensemble three paralleled running models, which incur a computational cost three times higher than that of ERM, as indicated in the time column. However, our method only requires a similar computational cost of ERM, which demonstrates the effectiveness of our method.
Since simply applying Lookahead bears overfitting issues, by adopting regularizations, our proposed AvgLookahead surpasses all these methods except for EoA. Note that by employing the same ensembling procedure as EoA, our method can achieve significantly better performance than EoA ($68.7\%$ vs. $68.0\%$).

% GFNet and,  GFNet~\cite{rao2021global} is based on the transformer while
Except for the backbone of Resnet-50, we also conduct validation of our proposed method using ResNeXt-50 backbone~\cite{yalniz2019billion}, where the architecture of ResNeXt-50 is a modified version of Resnet-50 and it is pre-trained on both ImageNet~\cite{deng2009imagenet} and a large weakly-semi-supervised dataset. This alternative backbone selection serves the purpose of verifying whether our method's effectiveness remains consistent across different architectures and whether it can still enhance performance when the baseline model is considerably stronger. As illustrated in the \cref{tab_sota}, the performance of the baseline (ERM) with a better ResNeXt-50 backbone improves significantly upon vanilla Resnet-50 (\ie, $68.6\%$ vs. $63.3\%$), while our proposed Lookahead algorithm can also improve on the stronger baselines ($72.5\%$ vs. $68.6\%$) and achieves comparable performance to EoA. Furthermore, when equipped with our proposed regularization operations, the performance can be further improved to $73.4\%$ and achieves SOTA performance.

\subsubsection{Semantic segmentation task}

\begin{figure*}[t]
  \centering
  \includegraphics[width=0.98\linewidth]{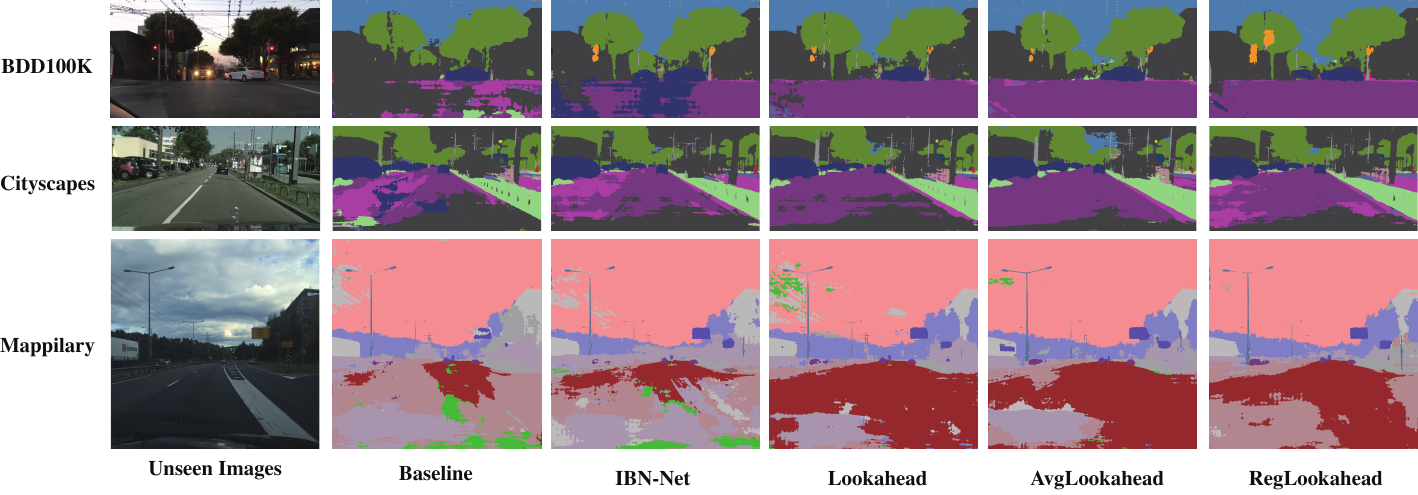}
  \caption{The semantic segmentation prediction comparison on three real-world datasets.}
\end{figure*}

\begin{table}[t]
  \begin{center}
    \caption{Performance comparison ($\%$) on the domain generalization semantic segmentation task (G$\rightarrow$(C, B, M)). The best performance is marked as \textbf{bold} and $^\dagger$ denotes that WildNet is trained with only source data.}
    \resizebox{1.0\columnwidth}{!}{
      \begin{tabular}{ l |c |  c c c | c c}
        \toprule
        Algorithm                               & Vennue  & C                & B                & M                & Avg.             \\%& G       \\
        \midrule
        Baseline                                & -       & $28.95$          & $25.14$          & $28.18$          & $27.42$          \\%& $73.45$ \\
        IBN-Net~\cite{pan2018two}               & ECCV-18 & $33.85$          & $27.48$          & $29.71$          & $30.35$          \\%&         \\
        RobustNet~\cite{choi2021robustnet}      & CVPR-20 & $36.58$          & $35.20$          & $40.33$          & $37.37$          \\%& $72.10$ \\
        PinTheMemory~\cite{kim2022pin}          & CVPR-22 & $41.00$          & $34.60$          & $37.40$          & $37.67$          \\
        % WildNet         & CVPR2022 & $44.62$ & $38.42$ & $46.09$ & $43.04$ & $71.20$ \\
        WildNet$^\dagger$~\cite{lee2022wildnet} & CVPR-22 & $40.10$          & $34.82$          & $39.38$          & $38.10$          \\%& $71.20$ \\
        \midrule
        Lookahead                               & Ours    & $40.27$          & $37.00$          & $41.41$          & $39.56$          \\%& $70.55$ \\
        AvgLookahead                            & Ours    & ${41.67}$ & ${39.10}$ & ${44.32}$ & ${41.70}$ \\%& $70.55$ \\
        RegLookahead                            & Ours    & $40.36$          & $38.00$          & $42.68$          & $40.35$          \\%& $70.55$ \\
        \midrule
        RobustNet+Lookahead                 & Ours    & $\textbf{42.15}$          & $\textbf{39.65}$          & $\textbf{45.70}$          & $\textbf{42.50}$          \\
        \bottomrule
      \end{tabular}
      \label{tab_segmentation}
    }
  \end{center}
  % \vspace{-15pt}
\end{table}

To further validate the effectiveness and wide application of our methods, we also conduct experiments on the semantic segmentation task of the domain generalization problem. The results are shown in \cref{tab_segmentation}. We compare several recent SOTA methods including IBN-Net~\cite{pan2018two} that embeds instance and batch normalization in one network, RobustNet~\cite{choi2021robustnet} that whitens features for domain-invariant feature learning, PinTheMemory~\cite{kim2022pin} that employs meta-learning algorithm, WildNet~\cite{lee2022wildnet} that augment feature with various styles. Note that, we only compare the WildNet model trained without ImageNet images for a fair comparison.
As demonstrated in the table, our proposed Lookahead can achieve the best performance among the three real-world datasets without changing the structure of the baseline like IBN-Net or adding extra losses, which makes it a more practical algorithm in the restricted areas.
Besides, although PinTheMemory employs the meta-learning training scheme, it adopts the similar training strategy of MLDG~\cite{li2018learning} that only employs a single trajectory with two steps and a small learning rate, which cannot help find the flat minima and achieve better generalization performance.
Although Lookahead can achieve better performance, it still suffers from the overfitting problem and by employing our proposed two different regularizations, the performance can be further boosted (\eg, $41.70\%$ vs. $39.56\%)$.
In addition, we also compare the performance of RobustNet~\cite{choi2021robustnet} with Lookahead, which can achieve better performance than simply training RobustNet~\cite{choi2021robustnet} and our methods on all three datasets, demonstrating the wide application of our method.

\subsection{Ablation Study}

In this section, we highlight the importance of two key elements (\ie, large learning rate and weight interpolation) in Lookahead and the regularizations in two variants of Lookahead by conducting an ablation study on the PACS dataset. The default hyperparameters are taken from the random search and are fixed for the ablation study.
As shown in \cref{tab_ablation}, when we only employ a large learning rate (\eg, $1e-4$ or $5e-4$), the performance degrades drastically (\eg, $75.40\%$ vs. $82.95\%$ with $5e-4$) since the model trained with a large learning rate has difficulty in convergence.
On the other hand, when we solely apply weight interpolation without the use of a large learning rate, the model's performance exhibits only marginal improvement (\eg, $84.68\%$ vs. $82.95\%$ with $\alpha=0.5$) but not significantly since the sparse interpolation of trained weight cannot produce an accurate estimation of flat minimum~\cite{cha2021swad}. By combining these two elements, the simple Lookahead training scheme with a longer inner loop can achieve significantly improved performance over ERM (\ie, $88.10\%$ vs. $82.95\%$ with $\eta=5e-4$, $\alpha=0.05$).
Since Lookahead still suffers from the overfitting problem, we propose two variants of it, namely RegLookahead and AvgLookahead, to further improve its generalization ability. As shown in \cref{tab_ablation}, by employing the averaging strategy or the l2 norm regularization, the training model can achieve better performance (\ie, $88.82\%$ and $88.33\%$ vs. $88.10\%$) than the vanilla Lookahead.
% Since AvgLookahead employs the average of the weights along the inner loop to perform regularization,   Besides, we utilize the averaged weight as the target weight of l2 norm regularization, which has little tendency to overfit the source data compared to the initial weight at each inner loop. Therefore, it achieves better performance ($88.33\%$ vs. $87.93\%$) during the test stage. Besides, we also compare the performance of the averaged model, which has a lower performance than the vanilla Lookahead since it has a strong regularization effect with weight averaging that lowers the performance of both Cartoon and Sketch, while our L2-norm regularization can reduce this problem by restricting a small range of the value of $\lambda$.

\begin{table}[t]
  \begin{center}
    \caption{The ablation study of our proposed Lookahead (LA), AvgLookahead (AvgLA), and RegLookahead (RegLA). The default inner and interpolation ratios are $\eta=5e-5$ and $\alpha=1$, respectively.}
    \resizebox{\columnwidth}{!}{
      \begin{tabular}{ l| c c| c c c c | c}
        \toprule
                                   & Large LR    & Interpolation & A       & C       & P       & S       & Avg.    \\
        \midrule
        ERM                        & $\eta=5e-5$ & $\alpha=1$    & $84.77$ & $73.72$ & $97.70$ & $75.63$ & $82.95$ \\
        \midrule
        $\eta$=$1e-4$              & \checkmark  &               & $78.65$ & $73.77$ & $95.43$ & $73.79$ & $80.41$ \\
        $\eta$=5e-4                & \checkmark  &               & $68.09$ & $71.70$ & $85.93$ & $75.70$ & $75.40$ \\
        $\alpha$=$0.05$            &             & \checkmark    & $85.35$ & $74.04$ & $97.83$ & $71.41$ & $82.16$ \\
        $\alpha$=0.5               &             & \checkmark    & $83.41$ & $81.24$ & $97.53$ & $76.56$ & $84.68$ \\
        $\eta$=5e-4, $\alpha$=0.05 & \checkmark  & \checkmark    & $88.43$ & $82.55$ & $97.03$ & $84.38$ & $88.10$ \\
        \midrule
        % RegLA (Init)              & \checkmark  & \checkmark & $89.41$ & $81.70$ & $97.25$ & $83.35$ & $87.93$ \\
        % w/ avg                    & \checkmark  & \checkmark & $89.35$ & $81.18$ & $97.42$ & $83.14$ & $87.77$ \\
        % \midrule
        AvgLA                      & \checkmark  & \checkmark    & $89.40$ & $83.97$ & $97.55$ & $84.35$ & $88.82$ \\
        RegLA                      & \checkmark  & \checkmark    & $89.20$ & $83.21$ & $97.73$ & $83.20$ & $88.33$ \\
        % $89.35_{\pm0.3}$ & $81.18_{\pm0.9}$ & $97.42_{\pm0.1}$ & $83.14_{\pm0.3}$ & $87.77_{\pm0.3}$
        % $89.41_{\pm0.3}$ & $81.70_{\pm0.6}$ & $97.25_{\pm0.1}$ & $83.35_{\pm0.5}$ & $87.93_{\pm0.3}$
        % Avg.        &            & \checkmark & \\
        \bottomrule
      \end{tabular}
      \label{tab_ablation}
    }
  \end{center}
  % \vspace{-20pt}
\end{table}

\subsection{Comparison to SAM and SWAD}

In this section, we compare the ability to identify flat minima of ERM, Lookahead with a large learning rate and two typical methods, \ie, SAM~\cite{foret2020sharpness}, SWAD~\cite{cha2021swad}.

% \textbf{Flatness comparison.} First, we plot the landscape flatness of the models trained by these methods in \cref{fig_sam_compare}. The flatness is calculated by perturbing model weights with randomly sampled filter-wise normalized noises~\cite{keskar2016large} and the results are averaged from 20 runs. Note that only relative changes are plotted (\ie, the minimum value of each line is set to equal.) As shown, the figure of ERM is the sharpest method while 

\begin{figure}[t]
  \centering
  % \vspace{-10pt}
  \subfloat[][ERM]{
    \hspace{-5pt}
    \includegraphics[width=0.25\linewidth]{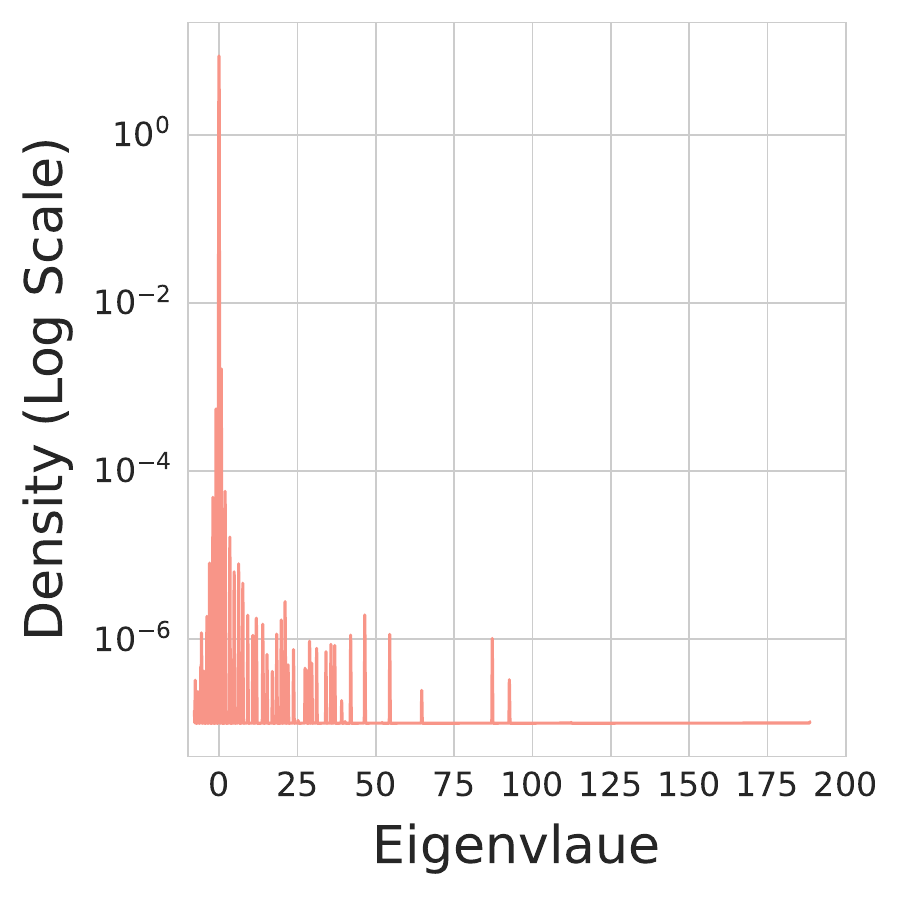}
    % \caption{The test accuracy} 
    \vspace{-5pt}
  }
  \subfloat[][SAM]{
    \hspace{-10pt}
    \includegraphics[width=0.25\linewidth]{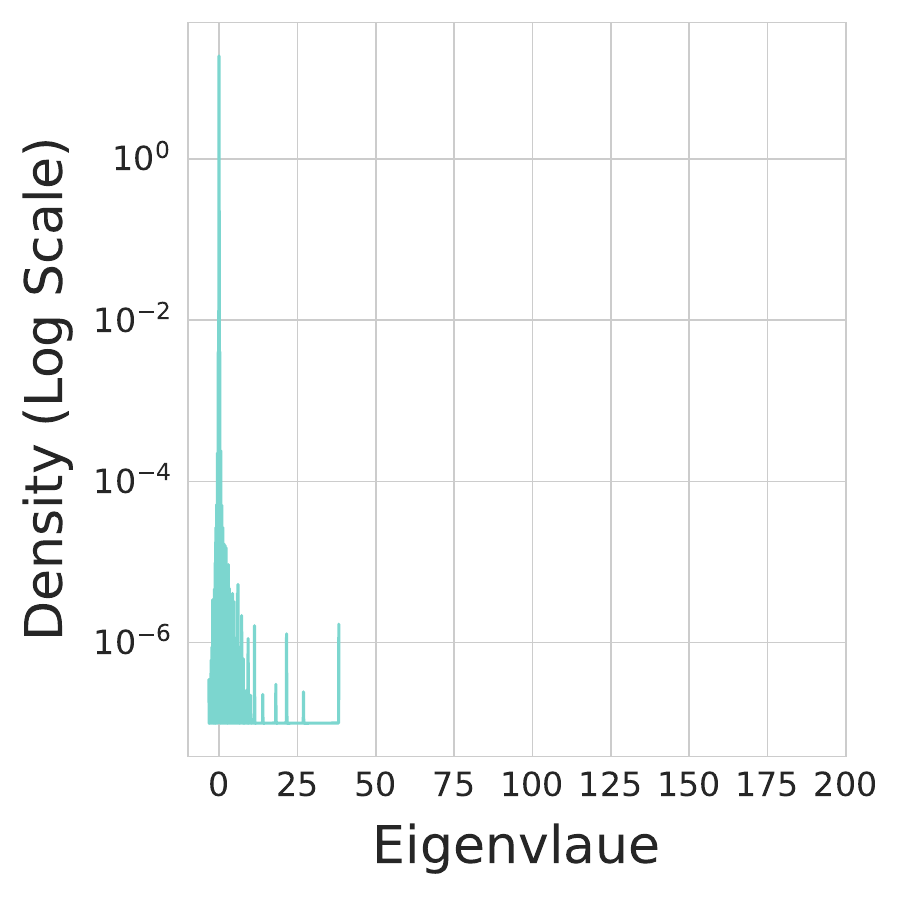}
    % \caption{The test accuracy}
    \vspace{-5pt}
  }
  \subfloat[][Lookahead]{
    \hspace{-10pt}
    \includegraphics[width=0.25\linewidth]{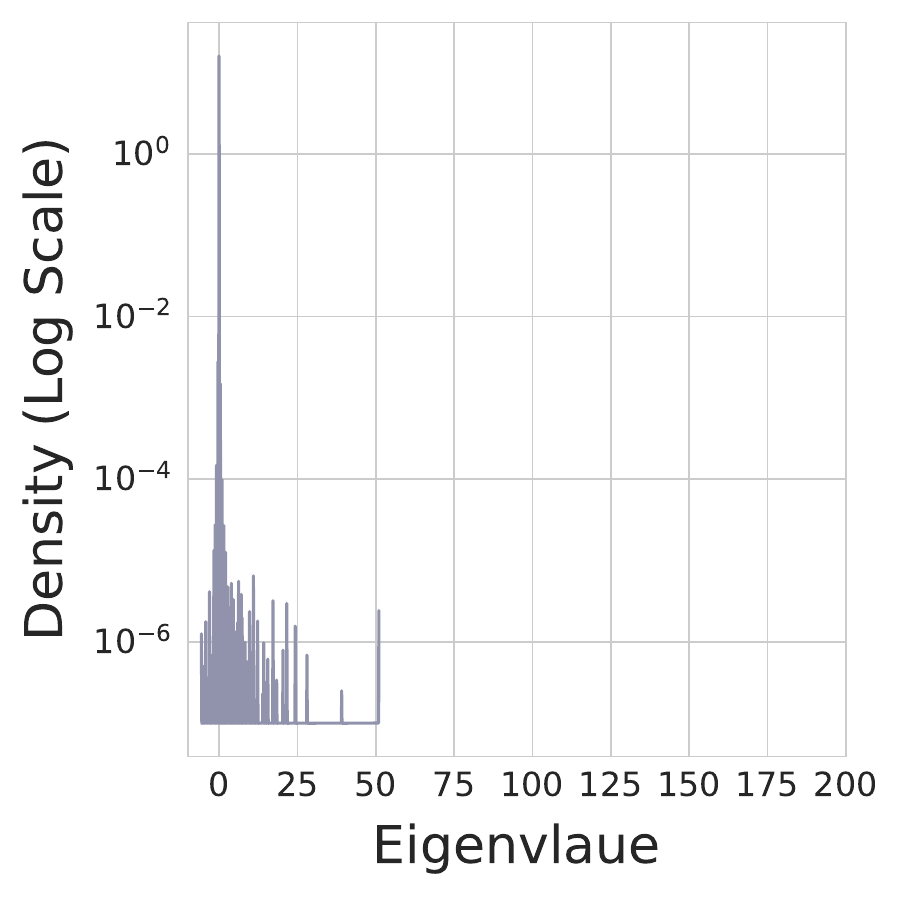}
    % \caption{The test accuracy}
    \vspace{-5pt}
  }
  \subfloat[][SWAD]{
    \hspace{-10pt}
    \includegraphics[width=0.25\linewidth]{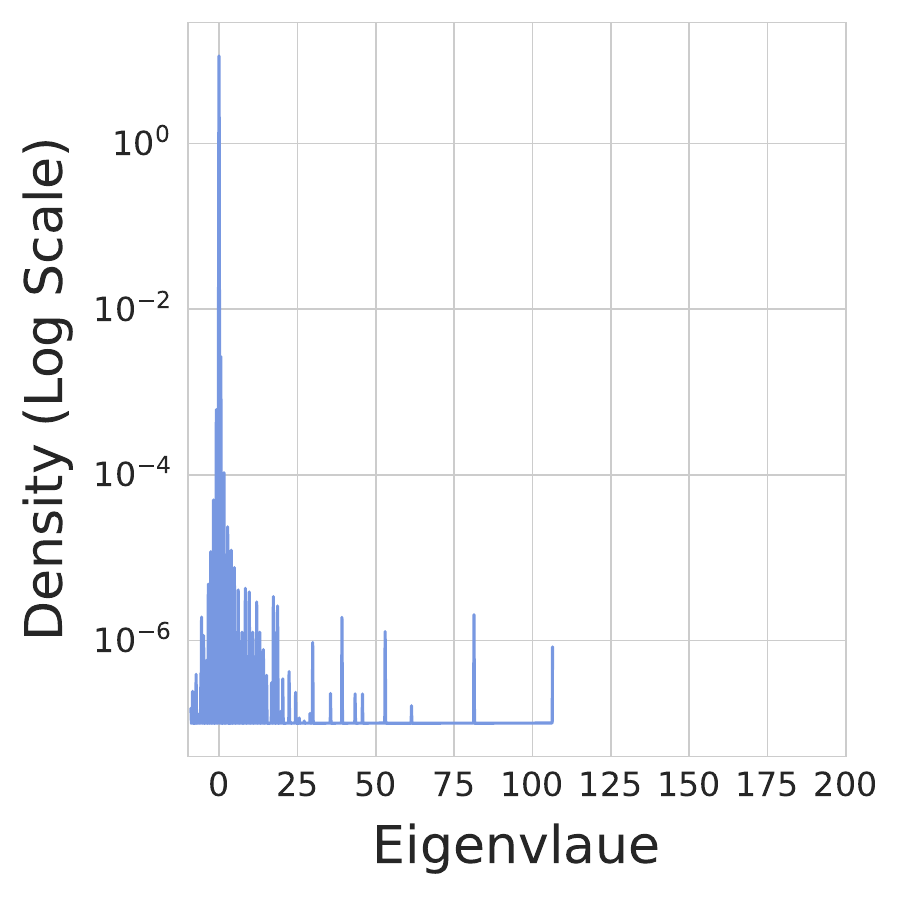}
    % \caption{The test accuracy}
    \vspace{-5pt}
  }
  \caption{The eigenvalue distributions obtained from models trained using ERM, SAM, SWAD, and Lookahead.}
  \label{fig_eigen_compare}
  % \vspace{-5pt}
\end{figure}

\textbf{Eigenvalues comparison of trained models.}  To assess whether these methods indeed identify flat minima, as the eigenvalues of the Hessian on the weights can indicate the sharpness of the current weight, we have plotted the distribution of eigenvalues for models trained using these techniques. As depicted in \cref{fig_eigen_compare}, the maximum eigenvalues for the different methods are as follows: $188$ for ERM, $38$ for SAM, $106$ for SWAD, and $51$ for Lookahead.
Comparing ERM and SWAD with SAM and Lookahead, it is evident that the latter two can identify flatter minima, where the eigenvalues are small and predominantly centered around `0'. This observation validates that SAM is indeed capable of discovering flat minima through the perturbation operation. Additionally, Lookahead, due to its utilization of a large learning rate, is also able to reduce the maximum eigenvalue to a lower value. However, SWAD does not exhibit a similar trend; although it does reduce the maximum eigenvalue from that of ERM (from $188$ to $106$), it still remains higher than the values for SAM and Lookahead ($37$ and $50$). Interestingly, SWAD demonstrates high performance, as indicated in \cref{tab_sota}. This suggests that a low maximum eigenvalue may not be the sole explanation for the high out-of-distribution performance.
Notably, although Lookahead and SWAD employ weight interpolation, Lookahead employs a large learning rate that guarantees the minimization of eigenvalues while the averaging weights of SWAD do not have this guarantee, which results in the minimization of eigenvalues as a by-product of weight averaging. Therefore, this may explain the reason why SWAD obtains high accuracy with relatively high eigenvalues.
%simply applying the averaging strategy might sometimes struggle to effectively reduce the maximum eigenvalue. 
This is further supported in \cite{kaddour2022flat}, where SWA even increased the maximum eigenvalue of GIN models. In contrast, Lookahead, through the use of a large learning rate and interpolation, achieves not only low eigenvalues but also superior performance.
%This dual benefit sets Lookahead apart, enabling it to strike a balance that leads to both flatter minima and improved performance.
%Besides, the distribution of eigenvalues computed with the ERM model is centered around `0' with a deviation approximately less than $250$, while the corresponding distribution of Lookahead has a deviation around $50$, which is significantly smaller and results in a flatter area that Lookahead finds.

% \textbf{Comparison of the gradient directions}

\begin{figure}[t]
  \centering
  % \vspace{-10pt}
  \subfloat[][Landscape on the validate set]{
    \includegraphics[width=0.45\linewidth]{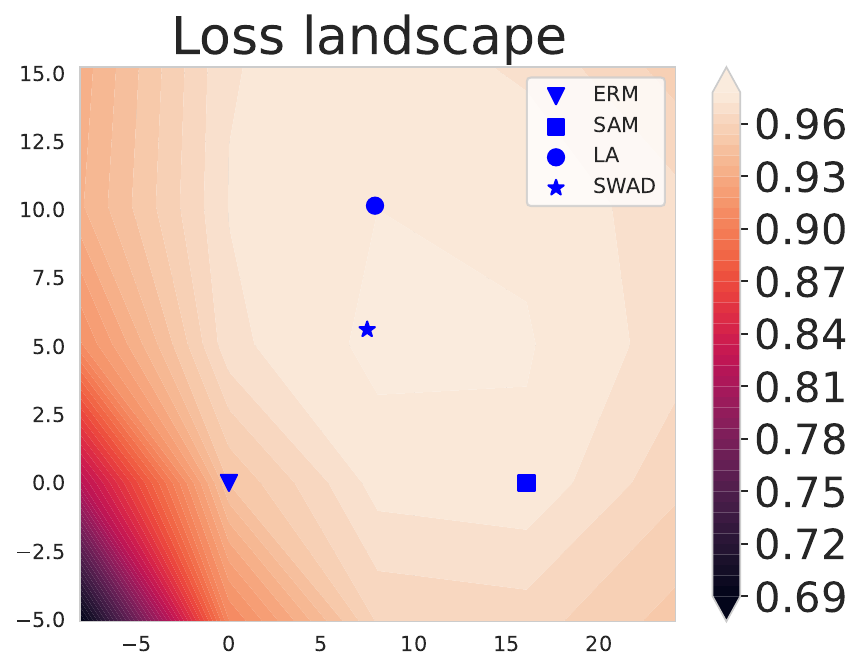}
    % \caption{The test accuracy}
    \vspace{-5pt}
  }
  \subfloat[][Landscape on the test set]{
    \includegraphics[width=0.45\linewidth]{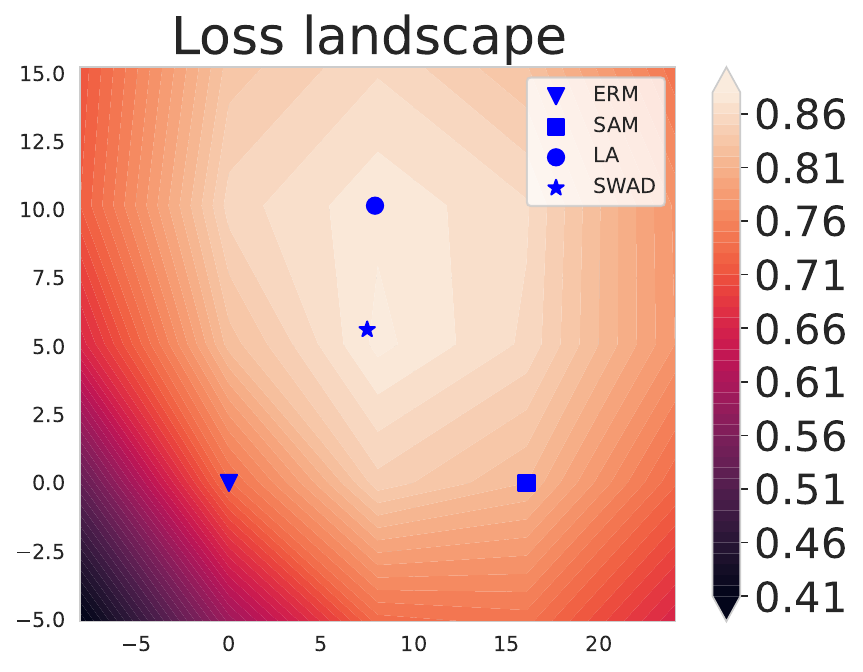}
    % \caption{The test accuracy}

    \vspace{-5pt}
  }
  \caption{The locations of ERM, SAM, SWAD, and Lookahead (LA) trained models on the projected loss landscapes.}
  \label{fig_landscape}
  \vspace{-5pt}
\end{figure}

\textbf{2D visualization of loss surfaces.} To assess the flatness of these methods, we conducted a comprehensive comparative analysis by evaluating the distribution of model weights on identical loss landscapes. These landscapes are visualized as 2D surfaces, obtained by projecting the model weights onto the space spanned by the weight vectors of ERM, SAM, and Lookahead (LA). The results, depicted in \cref{fig_landscape}, have provided us with valuable insights.
On the validation set, we observed that SAM, Lookahead (LA), and SWAD all occupied a region within the flat area of the landscape. This finding indicates their ability to identify flat minima, which is a crucial characteristic. In contrast, ERM's position primarily resides along the landscape's edge, confirming empirical observations from earlier studies~\cite{izmailov2018averaging,cha2021swad} that models trained using SGD have a tendency to remain near the edges of the loss landscape.
However, when faced with shifted data distributions, only SWAD and Lookahead manage to maintain their flatness, while both SAM and ERM move towards the landscape's edge. We hypothesize that SAM's training approach, which involves defining a radius that typically corresponds to a small region, limits its ability to extensively explore flatter areas. In contrast, SWA's strategy of weight averaging throughout the training trajectory enables the averaged weight to connect with every training weight, leading to a larger flat area.
For Lookahead, its use of a larger learning rate and extended inner loops empowers the fast weight to rapidly explore a broader region. This approach encourages the model to efficiently discover a more extensive and connected flat area through the interpolation operation. The combination of these factors contributes to the overall effectiveness of Lookahead in identifying flat minima.

\subsection{Further Analysis}
\label{sec_exp}

\begin{figure*}[t]
  \centering
  \subfloat[Miximum eigenvalue changes]{
    \includegraphics[width=0.22\linewidth]{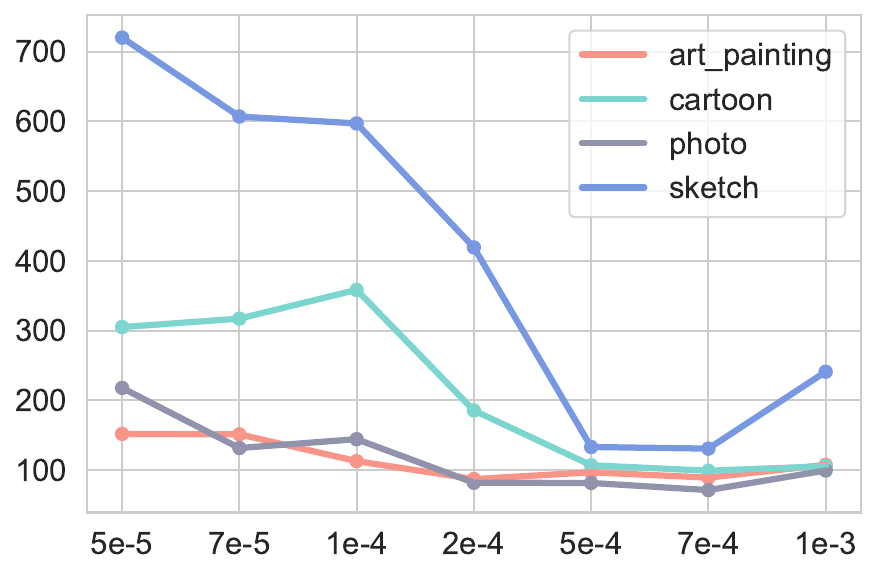}
    \label{fig_lr_eigenvalues}
  }
  \hspace{1pt}
  \subfloat[Accuracy and loss changes]{
    \includegraphics[width=0.22\linewidth]{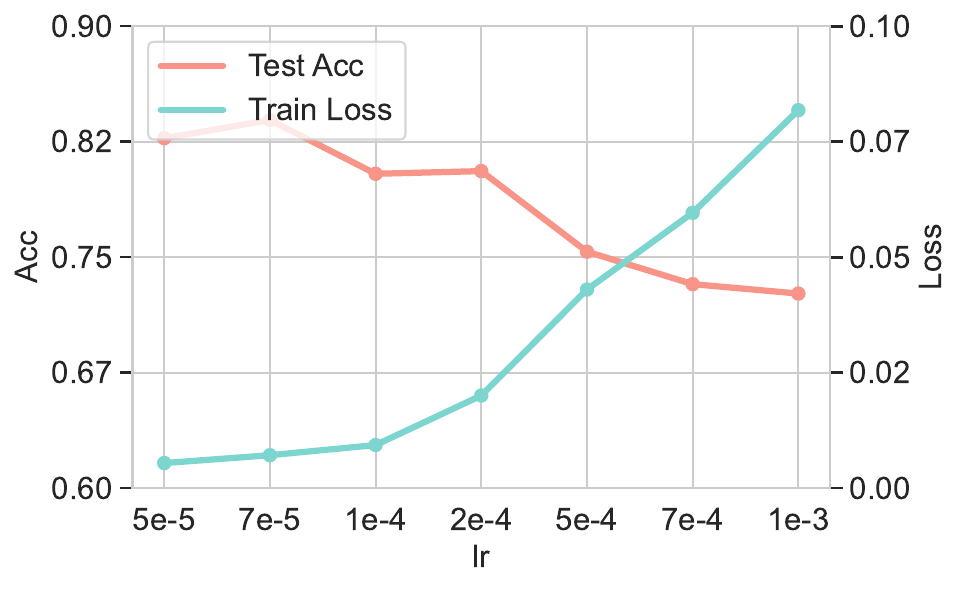}
    \label{fig_lr_acc}
  }
  \hspace{1pt}
  % \subfloat[Eigenvalue density of ERM]{
  %   \includegraphics[width=0.22\linewidth]{imgs/Hessian_ERM.pdf}
  %   \label{fig_eigenvalue_erm}
  % }
  % \hspace{1pt}
  % \subfloat[Eigenvalue density of LA]{
  %   \includegraphics[width=0.22\linewidth]{imgs/Hessian_Lookahead.pdf}
  %   \label{fig_eigenvalue_la}
  % }
  \subfloat[][The validation accuracy]{
    \includegraphics[width=0.23\linewidth]{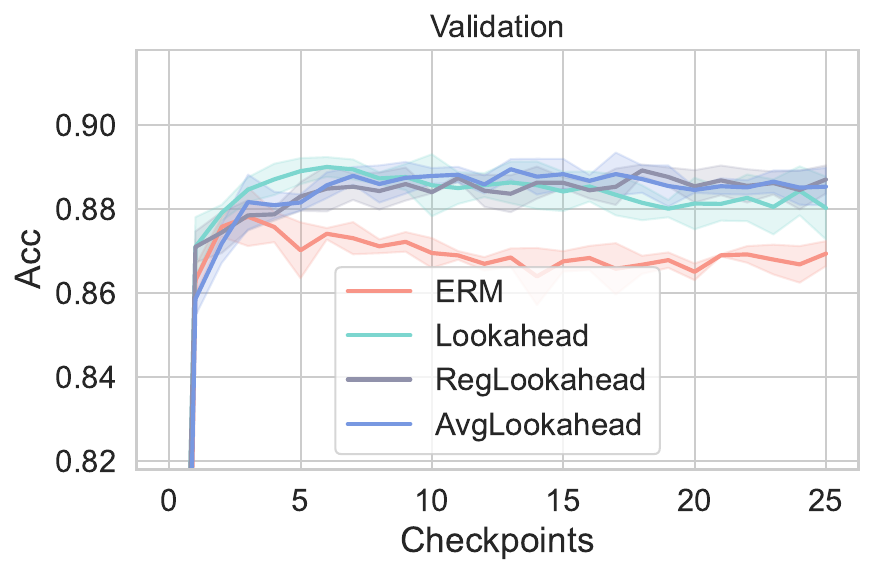}
    % \caption{The validation accuracy}
    \label{fig_stablility_a}
    \vspace{-5pt}
  }
  \subfloat[][The test accuracy]{
    \includegraphics[width=0.23\linewidth]{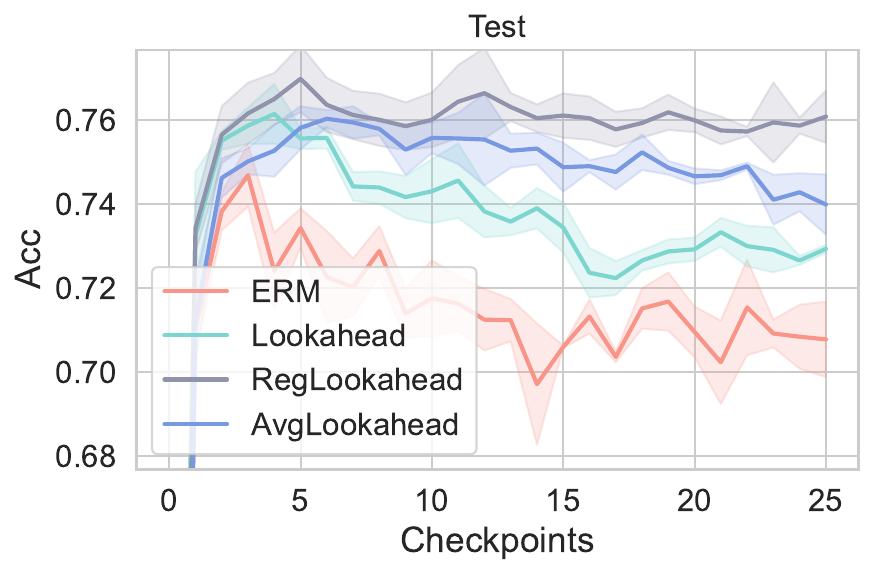}
    % \caption{The test accuracy}
    \label{fig_stablility_b}
    \vspace{-5pt}
  }
  % \vspace{-5pt}
  \caption{The maximum eigenvalue, accuracy, and loss changes according to the learning rate (a, b) and the comparison of validation and test accuracy changes during training (c, d).}
  % \vspace{-5pt}
  %   \caption{Different learning rates influence eigenvalues, accuracy and losses.}
  %   \label{fig_eigenvalues}
  %   \vspace{-5pt}
  % \end{figure}

  % \begin{figure}[t]
\end{figure*}

\textbf{The Eigenvalue changes according to different learning rates.} In \ref{sec_lr}, we posit that a larger learning rate increases the likelihood of the trained model staying within a flat region, which is inherently more robust to perturbations. Additionally, we hypothesize that in such cases, the corresponding maximum eigenvalues should exhibit a reduction. To empirically validate this theory, we computed and plotted the eigenvalues obtained from models trained with different learning rates on various target domains of the PACS dataset. The results, depicted in \cref{fig_lr_eigenvalues}, highlight an interesting pattern. When training with a larger learning rate, the eigenvalues across all target domains experience a substantial decrease, eventually plateauing at a certain value. This trend strongly indicates that a larger learning rate indeed facilitates the discovery of a flat area characterized by small eigenvalues.
Furthermore, we analyzed the average accuracy and loss changes on the PACS dataset as a function of the learning rate, illustrated in \cref{fig_lr_acc}. The observations are noteworthy: with a larger learning rate, there is a noticeable decline in accuracy, accompanied by an increase in training loss. This implies that the larger learning rate prevents the model from effectively fitting the data. However, it's crucial to note that the interpolation component within the Lookahead training scheme counteracts this issue, leading to an improvement in performance. This aspect highlights the power of the interpolation mechanism in mitigating the adverse effects of a large learning rate and ultimately enhancing the model's overall performance.

% how training loss, accuracy and eigenvalues change according to different learning rates. 

% \textbf{The distribution of eigenvalues.} 

\textbf{The overfitting problem and the effect of regularization.} We conduct an analysis to showcase the efficacy of AvgLookahead in regularizing the training process. In \cref{fig_stablility_a} and \cref{fig_stablility_b}, we plotted the training and test accuracy curves for AvgLookahead, Lookahead, and ERM.
As observed in \cref{fig_stablility_a}, both Lookahead and AvgLookahead exhibit a faster convergence rate, achieving better performance than ERM during the early stages of training. However, it becomes evident in \cref{fig_stablility_b} that Lookahead eventually starts overfitting the data, leading to a decline in test accuracy during the later stages.
In contrast, the application of regularization techniques in AvgLookahead mitigates the adverse effects on the training stage, ensuring that it doesn't detrimentally impact the test accuracy. As a result, AvgLookahead achieves a better and more stable performance compared to both Lookahead and ERM.
This highlights the substantial benefits of introducing regularization strategies in AvgLookahead, which effectively addresses the overfitting problem and contributes to more consistent and superior test accuracy.

% \begin{figure}[t]
%   \centering
%   \subfloat[][The validation accuracy]{
%     \includegraphics[width=0.46\linewidth]{imgs/VLCS-d2-val_New.pdf}
%     % \caption{The validation accuracy}
%     \label{fig_stablility_a}
%     \vspace{-5pt}
%   }
%   \subfloat[][The test accuracy]{
%     % \includegraphics[width=1\linewidth]{imgs/training_loss.pdf}
%     \includegraphics[width=0.46\linewidth]{imgs/VLCS-d2-test_New.pdf}
%     % \caption{The test accuracy}
%     \label{fig_stablility_b}
%     \vspace{-5pt}
%   }
%   \caption{The accuracy changes concerning checkpoints of ERM, Lookahead and AvgLookahead on the VLCS dataset.}
% \end{figure}

\textbf{Comparison to the variants of ERM with a large learning rate.}
To further investigate whether all the techniques mentioned in Lookahead take effect, we conduct a comparative analysis involving different variants of ERM. Specifically, we explored two variants of ERM: 1) training with a large learning rate: In this scenario, we initially trained the model with a large learning rate to identify a flat area and then gradually reduced the learning rate for optimization; 2) ensembling with varying frequencies: This approach involved ensembling models trained with a large learning rate at different frequencies. For example, we ensembled checkpoints every 15 steps along the trajectories, matching the inner loop length discovered during the search process.
The results are presented in \cref{tab_variants}. When trained with a large learning rate, the performance experienced a drastic decline ($82.95\%$ vs. $75.58\%$). Although adopting learning rate decay or ensembling improved performance to some extent, the results still fell significantly short of those achieved by Lookahead.
We hypothesize that while a large learning rate helps the model identify a flat area, this advantage becomes less effective when the learning rate is decayed. The model can quickly get trapped in sharp minima since the large learning rate prevents convergence, producing large gradients that push the model out of the flat area, leading to eventual convergence in a sharp area. In the case of ensembling, despite benefiting from the large learning rate, the models are still hard to converge (\ie, low performance (\eg, $83.71\%$) on the same domain (Photo) of the ImageNet pre-trained model), making it challenging for ensembling to produce models located in areas that are both low-loss and flat with better performance.

\begin{table}[t]
  \begin{center}
    \caption{Comparison to the variants of ERM trained with a large learning rate (\ie, $5e-4$). LR decay means the learning rate is decayed to $5e-5$ at the specific training step. Ensemble (X) means ensemble models every X step. }
    \resizebox{1\columnwidth}{!}{
      \begin{tabular}{ l |  c c c c | c}
        \toprule
        Algorithm               & A       & C       & P       & S       & Avg     \\
        \midrule
        ERM (lr=$5e-5$)         & $84.77$ & $73.72$ & $97.70$ & $75.63$ & $82.95$ \\
        ERM (lr=$5e-4$)         & $67.60$ & $69.56$ & $88.70$ & $76.46$ & $75.58$ \\
        \midrule
        LR decay  (2000)        & $75.11$ & $78.36$ & $92.14$ & $80.34$ & $81.49$ \\
        LR decay  (4000)        & $69.37$ & $78.04$ & $89.67$ & $76.81$ & $78.47$ \\
        \midrule
        Ensemble (5)            & $78.58$ & $79.05$ & $94.24$ & $80.50$ & $83.09$ \\
        Ensemble (10)           & $78.58$ & $80.65$ & $93.34$ & $81.30$ & $83.47$ \\
        Ensemble (15)           & $78.52$ & $80.60$ & $93.34$ & $81.36$ & $83.46$ \\
        SWAD~\cite{cha2021swad} & $78.04$ & $79.16$ & $93.71$ & $80.50$ & $82.85$ \\
        \midrule
        Lookahead               & $88.43$ & $82.55$ & $97.03$ & $84.38$ & $88.10$ \\
        \bottomrule
      \end{tabular}
      \label{tab_variants}
    }
  \end{center}
  % \vspace{-15pt}
\end{table}

\begin{table}[t]
  \begin{center}
    \caption{Performance comparison (\%) to the noise perturbated Lookahead with different strengths.}
    \resizebox{1\columnwidth}{!}{
      \begin{tabular}{ l |  c c c c | c}
        \toprule
        Algorithm          & A       & C       & P       & S       & Avg     \\
        \midrule
        ERM                & $84.77$ & $73.72$ & $97.70$ & $75.63$ & $82.95$ \\
        Lookahead          & $88.43$ & $82.55$ & $97.03$ & $84.38$ & $88.10$ \\
        + noise ($s=0.01$) & $87.61$ & $81.13$ & $96.93$ & $85.59$ & $87.82$ \\
        + noise ($s=0.05$) & $87.98$ & $83.69$ & $97.75$ & $81.71$ & $87.78$ \\
        + noise ($s=0.1$)  & $86.03$ & $84.06$ & $97.83$ & $83.91$ & $87.96$ \\
        + noise ($s=0.2$)  & $87.31$ & $82.84$ & $98.05$ & $82.82$ & $87.76$ \\
        \bottomrule
      \end{tabular}
      \label{tab_noise}
    }
  \end{center}
  % \vspace{-15pt}
\end{table}

\textbf{Comparison to the variant of Lookahead.} In \cref{sec_noise}, we argue that the noise cannot help the gradient estimation, as the mini-batch training strategy of SGD inherently contains noise. To verify this claim, we add noise during training with different strengths. The noise was generated by first sampling from a uniform distribution, resulting in $\hat{\epsilon}=\mathcal{U}(-1, 1)$. We then scaled this noise filter-wise, ensuring that it is comparable to the weights, a technique employed in~\cite{wen2018smoothout}. The final noise was obtained by applying a scaling factor, resulting in $\epsilon=s*\hat{\epsilon}$, where $s$ represents the noise strength.
As depicted in \cref{tab_noise}, the performance did not improve and, in fact, even experienced a slight drop when noise was applied at various strengths. This outcome aligns with the discussion in \cref{sec_noise}, where we mentioned that the mini-batch itself already contains inherent noise. Adding further noise during training could potentially exacerbate this situation, leading to a degradation in performance.

% \textbf{The effect of the length of inner loops.} We plot the effect of the inner loop length to validate that the longer inner loops produce a more accurate gradient estimation. As seen in \cref{fig_hyper_length}, a longer inner loop results in increased performance and then plateaus. Because the large step length takes the model weight away from the initial weight and if the weight is already far away, the estimated gradient does not change too much. As shown in \cref{fig_gradsim}, we plot the cosine similarity between the final gradient that we employ and the gradients calculated from each step. The similarity first increases gradually and then plateaus where it is high for the later weights. Therefore, in practice, a large training step is enough to obtain satisfactory performance. \re{TODO}

% \begin{figure}[h]
%   \subfloat[][learning rate]{
%     \includegraphics[width=0.45\linewidth]{imgs/ablation_inner_lr.pdf}
%     \label{fig_hyper_innerlr}
%   }
%   \subfloat[][interpolation ratio]{
%     \includegraphics[width=0.45\linewidth]{imgs/ablation_outer_lr.pdf}
%     \label{fig_hyper_outerLR}
%   }
%   \caption{The influence of the learning rate,  interpolation ratio, training steps and the grad similarity between different steps.}
%   \vspace{-5pt}
%   \label{fig_hypers}
% \end{figure}
% \textbf{The influence of average number and regularization effects.} To investigate how the number of weights to average and the strength of regularization affects model performance, we plot the performance of different average numbers and regularization strength in \cref{fig_hyper2}.

\begin{figure*}[t]
  \centering
  \subfloat[][learning rate]{
    \includegraphics[width=0.245\linewidth]{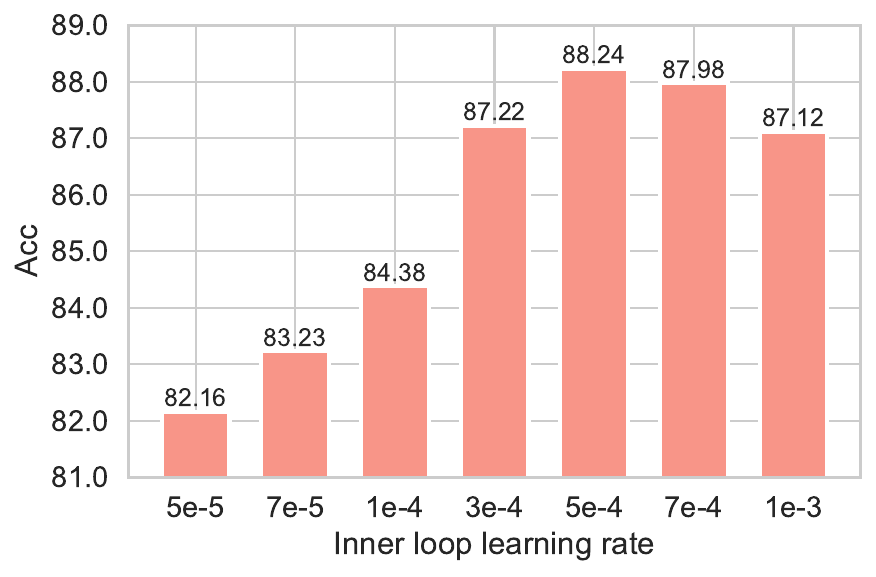}
    \label{fig_hyper_innerlr}
  }
  \subfloat[][interpolation ratio]{
    \hspace{-10pt}
    \includegraphics[width=0.245\linewidth]{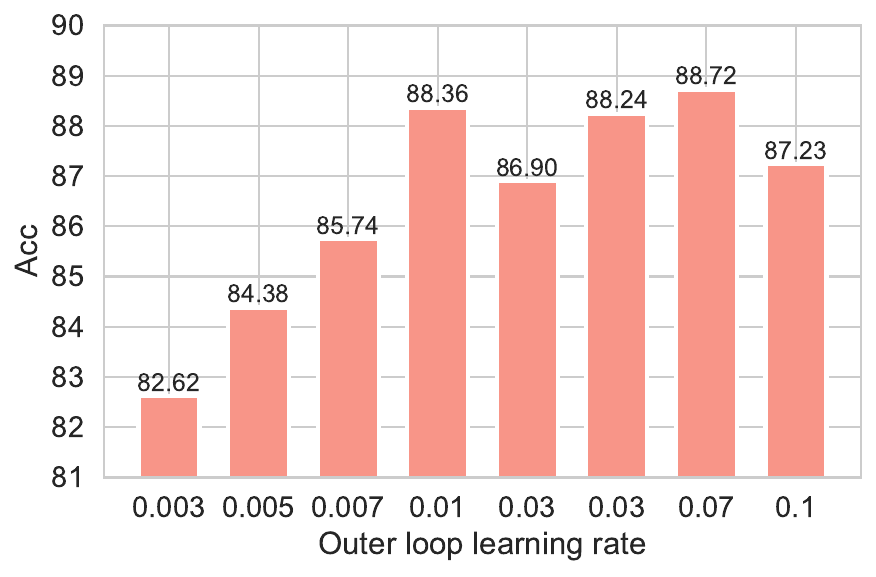}
    \label{fig_hyper_outerLR}
  }
  \subfloat[][Influence of training steps.]{
    \hspace{-10pt}
    \includegraphics[width=0.245\linewidth]{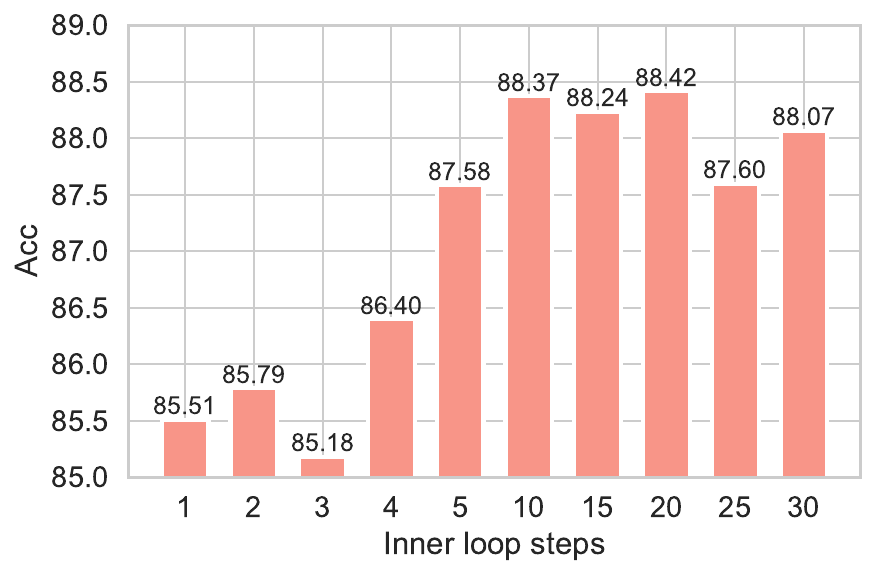}
    \vspace{-5pt}
    \label{fig_hyper_length}
  }
  \subfloat[][Flatness comparison.]{
    \hspace{-10pt}
    \includegraphics[width=0.22\linewidth]{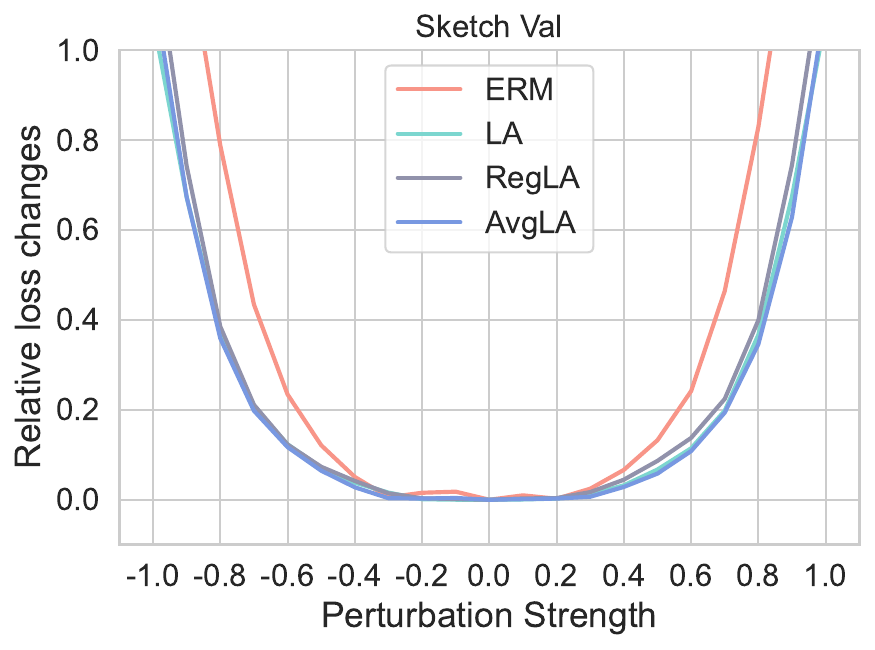}
    \vspace{-5pt}
    \label{fig_flat_val}
  }
  % \subfloat[][Gradient similarity.]{
  %   \includegraphics[width=0.25\linewidth]{imgs/grad_sim.pdf}
  %   \vspace{-5pt}
  %   \label{fig_gradsim}
  % }
  % \subfloat[][Flatness comparison.]{
  %   \includegraphics[width=0.25\linewidth]{imgs/compare_surface/LossSurfaceCompare_Sketch_test.pdf}
  %   \vspace{-5pt}
  %   \label{fig_flat_test}
  % }
  % \vspace{-7pt} 
  \caption{The influence of the learning rate, training length and interpolation ratio of (a, b, c) and the flatness comparison (c).}
  \vspace{-5pt}
  \label{fig_hyper2}
\end{figure*}

\textbf{Hyperparameters analysis in Lookahead.} In Lookahead, there are three crucial hyperparameters that contribute to the high performance. While we select these hyperparameters through grid-search on the validation set within the DomainBed benchmark, we also provide a visualization of their impact by changing one hyperparameter while keeping the others fixed. This allows us to understand how each hyperparameter influences the overall performance.
The influence of the learning rate is shown in \cref{fig_hyper2}. As previously discussed, a small learning rate leads to performance close to our proposed ERM baseline (as seen in \cref{tab_ablation} that when $\alpha=0.5, \eta=5e-5$, the performance is $84.68\%$, which is close to the performance of ERM $82.95\%$). Increasing the learning rate initially improves performance, but then it begins to decline. This trend indicates that a large learning rate indeed enhances generalization. However, an excessively large learning rate can hinder the model's ability to fit the training data well, causing a drop in performance.
Similar trends are observed with the interpolation ratio, as depicted in \cref{fig_hyper_outerLR}. A small interpolation ratio combined with a large learning rate results in poor performance, but increasing the interpolation ratio improves performance by facilitating weight interpolation, thereby enhancing the model's ability to fit the training data.
Finally, \cref{fig_hyper_length} illustrates that performance drops significantly with small step sizes, primarily due to the small interpolation ratio (as observed in \cref{tab_ablation}). A larger step size results in performance improvements, and the performance remains relatively stable even with larger step lengths. This is because a large step length moves the model weight away from the initial weight, and if the weight is already far from the initial point, the estimated gradient doesn't change dramatically. Hence, in practice, a moderately large training step is sufficient to achieve satisfactory performance.

\textbf{Loss landscape comparison.} We compare the relative loss changes of ERM, Lookahead, AvgLookahead, and RegLookahead after the model weight is perturbed with randomly sampled filter-wise normalized noises. The experiments are averaged from 20 runs.
As shown in \cref{fig_flat_val}, the ERM-trained model, in the absence of the Lookahead training scheme, demonstrates greater sensitivity to noisy perturbations.
In contrast, our proposed methods (Lookahead, AvgLookahead, and RegLookahead) exhibit robustness against these perturbations with better generalizability during testing. This robustness further highlights the effectiveness of our proposed methods in finding flatter minima and enhancing the model's ability to generalize well.

\section{Conclusion}
In this paper, different from previous flat minima searching algorithms that employ a small learning rate, which is limited in weight diversity, we provide a novel perspective to investigate the role of large learning rate in generating diverse weight and identifying flat minima for Domain Generalization. Then we introduce Lookahead to ease the optimization problem of solely adopting a large learning rate. This fast and slow weight interpolation strategy can effectively improve the convergence and also help identify flat minima when the fast weight is trained for more steps. Finally, to further enhance the ability of large learning rate, we design AvgLookahead and RegLookahead to regularize the weight by averaging the weights in the inner loop and or regularizing with accumulated history weights. Both Lookahead and its variants are proven to be effective on both domain generalization classification and semantic segmentation benchmarks with extensive experiments.

$\\$
% \noindent \textbf{Acknowledgment}: This work is supported by NSFC Program (62222604, 62206052, 62192783), China Postdoctoral Science Foundation Project (2023T160100), Jiangsu Natural Science Foundation Project (BK20210224), and CCF-Lenovo Bule Ocean Research Fund.

  {\small
    \bibliographystyle{IEEEtran}
    \bibliography{egbib}
  }

% \section{Biography Section}
% If you have an EPS/PDF photo (graphicx package needed), extra braces are
%  needed around the contents of the optional argument to biography to prevent
%  the LaTeX parser from getting confused when it sees the complicated
%  $\backslash${\tt{includegraphics}} command within an optional argument. (You can create
%  your own custom macro containing the $\backslash${\tt{includegraphics}} command to make things
%  simpler here.)

% \vspace{11pt}

% \bf{If you include a photo:}\vspace{-33pt}
% \begin{IEEEbiography}[{\includegraphics[width=1in,height=1.25in,clip,keepaspectratio]{fig1}}]{Michael Shell}
% Use $\backslash${\tt{begin\{IEEEbiography\}}} and then for the 1st argument use $\backslash${\tt{includegraphics}} to declare and link the author photo.
% Use the author name as the 3rd argument followed by the biography text.
% \end{IEEEbiography}

% \vspace{11pt}

% \bf{If you will not include a photo:}\vspace{-33pt}
% \begin{IEEEbiographynophoto}{John Doe}
% Use $\backslash${\tt{begin\{IEEEbiographynophoto\}}} and the author name as the argument followed by the biography text.
% \end{IEEEbiographynophoto}

\onecolumn

\section{Deductions}
\label{learning_rate}

\subsection{The deduction of the relationship of two learning rates and $h_{\max}$ in Lookahead}

In this section, we only consider the vanilla Lookahead. The expectation of the updating trajectory of Lookahead can be calculated as
$$
  \begin{aligned}
    \mathbb{E}\left[\boldsymbol{\phi}_{t+1}\right]      & =(1-\alpha)  \mathbb{E}\left[\boldsymbol{\phi}_t\right]+\alpha  \mathbb{E}\left[\theta_{t,k}\right]                                                                                                       \\
                                                        & =(1-\alpha) \mathbb{E}\left[\boldsymbol{\phi}_t\right]+\alpha \left(\mathbf{\mathbf{I}}-\eta \mathbf{H})^k \mathbb{E}\left[\boldsymbol{\phi}_t\right]\right.                                            \\
                                                        & =\left[1-\alpha+\alpha\left(\mathbf{\mathbf{I}}-\eta \mathbf{H}\right)^k \right] \mathbb{E}\left[\boldsymbol{\phi}_t\right]                                                                                      \\
    \mathbb{E}\left[\boldsymbol{\phi}_{t+1}{ }^2\right] & =\mathbb{E}\left\{\boldsymbol{\phi}_t^{\top} \left[1-\alpha+\alpha(\mathbf{\mathbf{I}}-\eta \mathbf{H})^k\right]^2  \boldsymbol{\phi}_t\right\}                                                                  \\
                                                        & =\mathbb{E} \left\{ \boldsymbol{\phi}_t^{\top} \left[(1-\alpha)^2+2 \alpha (1-\alpha) (\mathbf{\mathbf{I}}-\eta \mathbf{H})^k+\alpha^2 (\mathbf{\mathbf{I}}-\eta \mathbf{H})^{2 k}\right]  \boldsymbol{\phi}_t \right\}
  \end{aligned}
$$
To ensure the convergence, instead of simply require that $ \left[(1-\alpha)^2+2 \alpha (1-\alpha) (1-\eta h_{\max})^k+\alpha^2 (1-\eta h_{\max})^{2 k}\right] < 1$, we also require $\alpha^2(1-\eta h_{\max})^{2k} < 1$ to ensure the convergence. To meet this requirement, we only need to meet $\alpha^2 (1-\eta h_{\max})^{2 k} < 1$, which gives $\frac{1}{\eta} - \frac{1}{\eta}\left(\frac{1}{\alpha}\right)^{1 / k} < h_{\max} < \frac{1}{\eta}\left(\frac{1}{\alpha}\right)^{1 / k}+\frac{1}{\eta}
$.

\subsection{The deduction of Variance Reduction} 
\label{variance}

Let $\boldsymbol{\phi}_{t}$ as the weight of iteration $t$ and $\mathbf{M} = (\mathbf{\mathbf{\mathbf{I}}} - \eta \mathbf{H})$. The updating rule of SGD is $\boldsymbol{\phi}_{t+1}=(\mathbf{\mathbf{I}}-\eta\mathbf{H})\boldsymbol{\phi}_{t}$. SGD has the following trajectories:
$$
  \begin{aligned}
     & \mathbb{E}\left[\boldsymbol{\phi}_{t+1}\right]=\mathbf{M}  \mathbb{E}\left[\boldsymbol{\phi}_t\right]                                 \\
     & \mathbb{V}\left[\boldsymbol{\phi}_{t+1}\right]=\mathbf{M}^2  \mathbb{V}\left[\boldsymbol{\phi}_t\right]+\eta^2  \mathbf{H}^2 \Sigma
  \end{aligned}
$$

Now we let $\boldsymbol{\phi}^{t}$ as the slow weight of iteration $t$ in outer loops and $\theta_{t,i}$ as the fast weight of $t$-th iteration in the outer loop and $i$-th iteration in the inner loop. Its trajectories are as follows:
$$
  \begin{aligned}
    \mathbb{E}\left[\boldsymbol{\phi}_{t+1}\right] & =(1-\alpha)  \mathbb{E}\left[\boldsymbol{\phi}_t\right]+\alpha  \sum_{i=0}^{k-1} \beta_i \mathbb{E}\left[\boldsymbol{\theta}_{t, i}\right]             \\
                                                   & =(1-\alpha)  \mathbb{E}\left[\boldsymbol{\phi}_t\right]+\alpha  \sum_{i=0}^{k-1} \beta_i M^i  \mathbb{E}\left[\boldsymbol{\phi}_t\right]               \\
                                                   & =\left[1-\alpha+\alpha \sum_{i=0}^{k-1} \beta_i\left(\mathbf{\mathbf{I}}-\eta \mathbf{H}\right)^i\right]  \mathbb{E}\left[\boldsymbol{\phi}_t\right]
  \end{aligned}
$$
We first obtain several components that are required for calculating $\mathbb{V}[\boldsymbol{\phi}_{t+1}]$:
$$
  \begin{aligned}
    \mathbb{\mathbb{V}}\left[\boldsymbol{\theta}_{t, k}\right]                            & = \left[\mathbf{M}^{2k}  \mathbb{V}\left[\boldsymbol{\phi}_t\right]+\sum_{i=0}^{k-1} \mathbf{M}^{2 i}  \eta^2 \mathbf{H}^2  \Sigma\right]                                                                                                                                                                                                                                                                                                                           \\
    \operatorname{Cov}\left(\boldsymbol{\theta}_{t, j}, \boldsymbol{\theta}_{t, i}\right) & =\mathbf{M}^{j-i}  \mathbb{V}\left[\boldsymbol{\theta}_i\right]                                                                                                                                                                                                                                                                                                                                                                                                       \\
                                                                                          & =\mathbf{M}^{j-i} \left[\mathbf{M}^{2 i}  \mathbb{V}\left[\boldsymbol{\phi}_t\right]+\sum_{l=0}^{i-1}\mathbf{M}^{2 l}  \eta^2  \mathbf{H}^2 \Sigma\right]                                                                                                                                                                                                                                                                                                           \\
                                                                                          & =\mathbf{M}^{j + i}  \mathbb{V}\left[\boldsymbol{\phi}_t\right]+\sum_{l=0}^{i-1}\mathbf{M}^{2l + j - i} \eta^2 \mathbf{H}^2 \Sigma                                                                                                                                                                                                                                                                                                                                  \\                                                                                     \mathbb{V}\left(\sum_{i=0}^{k-1} \beta_i  \boldsymbol{\theta}_{t, i}\right)  &=\sum_{i=0}^{k-1} \beta_i^2  \mathbb{V}\left(\boldsymbol{\theta}_{t, i}\right)+2 \sum_{i=0}^{k-1}\sum_{j=0}^{i-1} \beta_i \beta_j \operatorname{cov}\left(\boldsymbol{\theta}_{t, i}, \boldsymbol{\theta}_{t j}\right) \\
                                                                                          & =\sum_{i=0}^{k-1}  \beta_i^2 \left[\mathbf{M}^{2 i}  \mathbb{V}\left[\boldsymbol{\phi}_t\right]+\sum_{l=0}^{i-1} \mathbf{M}^{2l} \eta^2 \mathbf{H}^2  \Sigma\right]                                                                                                                                                                                                                                                                                                 \\
                                                                                          & +2 \sum_{i=0}^{k-1}\sum_{j=0}^{i-1} \beta_i \beta_j\left[\mathbf{M}^{i+j}  \mathbb{V}\left[\boldsymbol{\phi}_t\right]+\sum_{i=0}^{i-1}\mathbf{M}^{2l+j-i}  \eta^2  \mathbf{H}^2  \Sigma\right]                                                                                                                                                                                                                                                                      \\
                                                                                          & =\left[\sum_{i=0}^{k-1} \beta_i^2 \mathbf{M}^{2i}+2 \sum_{i=0}^{k-1}\sum_{j=0}^{i-1} \beta_i \beta_j\mathbf{M}^{j+i}\right] \mathbb{V}\left[\boldsymbol{\phi}_t\right]         +\eta^2  \mathbf{H}^2  \Sigma\left[\sum_{i=0}^{k-1} \beta_i^2 \sum_{i=0}^{i-1} \mathbf{M}^{2 l}+2 \sum_{i=0}^{k-1}\sum_{j=0}^{i-1} \beta_i \beta_j \sum_{l=0}^{i-1} \mathbf{M}^{2 l + j - i}\right]                                                                                  \\
                                                                                          & = \left[\sum_{i=0}^{k-1} \beta_i^2 \mathbf{M}^{2i}+2 \sum_{i=0}^{k-1}\sum_{j=0}^{i-1} \beta_i \beta_j\mathbf{M}^{j+i}\right] \mathbb{V}\left[\boldsymbol{\phi}_t\right]         +\eta^2  \mathbf{H}^2  \Sigma \left[\sum_{i=0}^{k-1} \beta_i^2  \frac{\mathbf{I}-\mathbf{M}^{2i}}{\mathbf{I}-\mathbf{M}^2}+2  \sum_{i=0}^{k-1} \sum_{j=0}^{i-1} \beta_i  \beta_j  \frac{\mathbf{M}^{j-i} \left[\mathbf{I}-\mathbf{M}^{2 j}\right]}{\mathbf{I}-\mathbf{M}^2} \right] \\
                                                                                          & = \left[\sum_{i=0}^{k-1} \beta_i^2 \mathbf{M}^{2i}+2 \sum_{i=0}^{k-1}\sum_{j=0}^{i-1} \beta_i \beta_j\mathbf{M}^{j+i}\right] \mathbb{V}\left[\boldsymbol{\phi}_t\right]   + \frac{\eta^2  \mathbf{H}^2  \Sigma}{\mathbf{I}-\mathbf{M}^2} \left[\sum_{i=0}^{k-1}  \beta_i^2 \left(\mathbf{I}-\mathbf{M}^{2i}\right)+2  \sum_{i=0}^{k-1} \sum_{j=0}^{i-1} \beta_i \beta_j \mathbf{M}^{i-j} (\mathbf{I}-\mathbf{M}^{2j})\right]                                        \\
                                                                                          & =\left[\sum_{i=0}^{k-1} \beta_i^2 \mathbf{M}^{2i}+2 \sum_{i=0}^{k-1}\sum_{j=0}^{i-1} \beta_i \beta_j\mathbf{M}^{j+i}\right] \mathbb{V}\left[\boldsymbol{\phi}_t\right]   +  \frac{\eta^2  \mathbf{H}^2  \Sigma}{\mathbf{I}-\mathbf{M}^2} Y
  \end{aligned}
$$

Then we can obtain  $\mathbb{V}[\boldsymbol{\phi}_{t+1}]$ as follows:
$$
  \begin{aligned}
    \mathbb{V}\left[\boldsymbol{\phi}_{t+1}\right] & =\mathbb{V}\left[(1-\alpha) \boldsymbol{\phi}_t+\alpha  \sum_{i=1}^k \beta_i \boldsymbol{\theta}_{t i}\right]                                                                                                                                                        \\
                                                   & =(1-\alpha)^2 \mathbb{V}\left(\boldsymbol{\phi}_t\right)+2 \alpha (1-\alpha)  \operatorname{Cov}\left(\boldsymbol{\phi}_t, \sum_{i=0}^{k-1} \beta_i \boldsymbol{\theta}_{t,i}\right) +\alpha^2  \mathbb{V}\left(\sum_{i=1}^{k-1} \boldsymbol{\theta}_{t, i}\right)   \\
                                                   & =(1-\alpha)^2  \mathbb{V}\left(\boldsymbol{\phi}_t\right)+2 \alpha (1-\alpha) \sum_{i=0}^{k-1} \beta_i  \operatorname{Cov}\left(\boldsymbol{\phi}_{t,} \boldsymbol{\theta}_{t i}\right)+\alpha^2  \mathbb{V}\left(\sum_{i=0}^{k-1} \boldsymbol{\theta}_{t, i}\right) \\
                                                   & =(1-\alpha)^2  \mathbb{V}\left(\boldsymbol{\phi}_t\right)+2 \alpha(1-\alpha) \sum_{i=0}^{k-1} \beta_i\mathbf{M}^i  \mathbb{V}\left[\boldsymbol{\phi}_t\right]+\alpha^2 \mathbb{V}\left(\sum_{i=0}^{k-1} \beta_i  \boldsymbol{\theta}_{t i}\right)                    \\
                                                   & =\left[(1-\alpha)^2+2 \alpha (1-\alpha)  \sum_{i=0}^{k-1} \beta_i\mathbf{M}^i \right] \mathbb{V}\left(\boldsymbol{\phi}_t\right)+\alpha^2 \mathbb{V}\left(\sum_{i=1}^{k-1} \beta_i \boldsymbol{\theta}_{t, i}\right)                                                 \\
  \end{aligned}
$$

$$
  \begin{aligned}
    \mathbb{V}\left[\boldsymbol{\phi}_{t+1}\right] & = \left[(1-\alpha)^2+2 \alpha (1-\alpha)  \sum_{i=0}^{k-1} \beta_i\mathbf{M}^i + \alpha^2\sum_{i=0}^ {k-1} \beta_i^2 \mathbf{M}^{2i}+2 \alpha^2\sum_{i=0}^{k-1}\sum_{j=0}^{i-1} \beta_i \beta_j\mathbf{M}^{j+i}\right] \mathbb{V}\left[\boldsymbol{\phi}_t\right]   +  \frac{\alpha^2\eta^2  \mathbf{H}^2  \Sigma}{\mathbf{I}-\mathbf{M}^2}  Y \\
                                                   & = \left[(1-\alpha)+\alpha  \sum_{i=0}^{k-1}  \beta_i\mathbf{M}^i\right]^2  \mathbb{V}\left[\boldsymbol{\phi}_t\right] +  \frac{\alpha^2 \eta^2  \mathbf{H}^2  \Sigma}{\mathbf{I}-\mathbf{M}^2}  Y                                                                                                                                              \\
  \end{aligned}
$$

When $\mathbb{V}_{\boldsymbol{\phi}_{t+1}}$ is optimal, that is $\mathbb{V}_{\boldsymbol{\phi}_{t+1}}=\mathbb{V}_{\boldsymbol{\phi}_{t}}=\mathbb{V}^*_{\text{AvgLA}}$, then we can obtain $\mathbb{V}^*_{\text{AvgLA}}$:
$$
  \begin{aligned}
    \mathbb{V}^*_{\text{AvgLA}} & = \frac{Y}{\mathbf{I} - \left[(1-\alpha)\mathbf{I}+\alpha  \sum_{i=0}^{k-1}  \beta_i\mathbf{M}^i\right]^2 } \frac{\alpha^2 \eta^2  \mathbf{H}^2  \Sigma}{\mathbf{I}-\mathbf{M}^2}
  \end{aligned}
$$

To compare with
$V_{ERM}^*   =\frac{\eta^2 \mathbf{H}^2 \Sigma^2}{\mathbf{\mathbf{I}}-\mathbf{M}^2}$ and $V_{L A}^*=\frac{\alpha^2\left(\mathbf{\mathbf{I}}-\mathbf{M}^{2 k}\right)}{\alpha^2\left(\mathbf{\mathbf{I}}-\mathbf{M}^{2 k}\right)+2 \alpha(1-\alpha)\left(\mathbf{\mathbf{I}}-\mathbf{M}^k\right)} V_{ERM}^*$, we need to compare the numerator and denominator:

$$
  \begin{aligned}
    Y & = \sum_{i=0}^{k-1}  \beta_i^2 \left(\mathbf{I}-\mathbf{M}^{2i}\right)+2  \sum_{i=0}^{k-1} \sum_{j=0}^{i-1} \beta_i \beta_j \mathbf{M}^{i-j} (\mathbf{I}-\mathbf{M}^{2j}) \\
      & = \sum_{i=0}^{k-1}  \beta_i^2 \left(\mathbf{I}-\mathbf{M}^{2i}\right)+2  \sum_{i=0}^{k-1} \sum_{j=0}^{i-1} \beta_i \beta_j \mathbf{M}^{i-j} (\mathbf{I}-\mathbf{M}^{2j}) \\
      & = \sum_{i=0}^{k-1}  \beta_i^2 \left(\mathbf{I}-\mathbf{M}^{2i}\right)+2  \sum_{i=0}^{k-1} \sum_{j=0}^{i-1} \beta_i \beta_j (\mathbf{M}^{i-j} - \mathbf{M}^{i+j})         \\
      & \le \sum_{i=0}^{k-1}  \beta_i^2 \left(\mathbf{I}-\mathbf{M}^{2i}\right)+2  \sum_{i=0}^{k-1} \sum_{j=0}^{i-1} \beta_i \beta_j (\mathbf{I} - \mathbf{M}^{i+j})             \\
      & = \sum_{i=0}^{k-1} \sum_{j=0}^{k-1} \beta_i\beta_j(\mathbf{I} - \mathbf{M}^{i+j})                                                                                        \\
      & = \mathbf{I} - \sum_{i=0}^{k-1} \sum_{j=0}^{k-1} \beta_i\beta_j \mathbf{M}^i \mathbf{M}^j                                                                                \\
      & = \mathbf{I} - (\sum_{i=0}^{k-1} \beta_i \mathbf{M}^i)^2
  \end{aligned}
$$

% \clearpage

let $\beta_i = \frac{c^i}{\sum^{k-1}_{i=0}c^i}$ , Since $\mathbf{0} \le max(\mathbf{M}) \le \mathbf{I}$ :
$$
  \begin{aligned}
    \frac{\sum_{i=0}^{k-1} \beta_i \mathbf{M}^i}{\mathbf{M}^{k}} & = \frac{\sum_{i=0}^{k-1}c^i \mathbf{M}^i}{\mathbf{M}^{k} \sum_{i=0}^{k-1}c^i } = \frac{\sum_{i=0}^{k-1}\frac{c^i}{\mathbf{M}^{k-i}}}{\sum_{i=0}^{k-1}c^i} \ge \mathbf{I} \\
    \sum_{i=0}^{k-1} \beta_i \mathbf{M}^i                        & \ge \mathbf{M}^k
  \end{aligned}
$$
Therefore, $Y \le \mathbf{I} - \mathbf{M}^{2k}$. Besides :

$$
  \begin{aligned}
        & \mathbf{I} - \left[(1-\alpha)\mathbf{I}+\alpha  \sum_{i=0}^{k-1}  \beta_i \mathbf{M}^i\right]^2 \\
    \ge & \mathbf{I} - \left[(1-\alpha)\mathbf{I}+\alpha  \sum_{i=0}^{k-1}  \beta_i \mathbf{M}^k\right]^2 \\
    =   & \mathbf{I} - \left[(1-\alpha)\mathbf{I}+\alpha  \mathbf{M}^k\right]^2
  \end{aligned}
$$

Therefore,
$$
  \begin{aligned}
    \mathbb{V}^*_{\text{WLA}} & = \frac{\alpha^2 Y}{\mathbf{I} - \left[(1-\alpha)\mathbf{I}+\alpha  \sum_{i=0}^{k-1}  \beta_i\mathbf{M}^i\right]^2 } \frac{\eta^2  \mathbf{H}^2  \Sigma}{\mathbf{I}-\mathbf{M}^2}    \\
                              & \le \frac{\alpha^2 (\mathbf{I}-\mathbf{M}^2)}{\mathbf{I} - \left[(1-\alpha)\mathbf{I}+\alpha  \mathbf{M}^{k}\right]^2 } \frac{\eta^2  \mathbf{H}^2  \Sigma}{\mathbf{I}-\mathbf{M}^2} \\
                              & \le \mathbb{V}^*_{\text{LA}} \le \mathbb{V}^*_{\text{ERM}}                                                                                                                             \\
  \end{aligned}
$$

\vfill

\end{document}